\documentclass[10pt,journal,compsoc]{IEEEtran}
\usepackage{times}
\usepackage{epsfig}
\usepackage{graphicx}
\usepackage{amsmath}
\usepackage{amssymb}
\usepackage{pifont}
\usepackage{xcolor}
\usepackage{framed}
\usepackage{amssymb}
\usepackage{latexsym}
\usepackage{epsfig}
\usepackage{float}
\usepackage{booktabs,multirow, multicol}
\usepackage{rotating}
\usepackage{epstopdf}
\usepackage{graphicx}
\usepackage{hyperref}

\usepackage[normalem]{ulem}
\usepackage{hhline}
\useunder{\uline}{\ul}{}

\usepackage{array, longtable}
\usepackage{wrapfig}
\usepackage{enumitem}
\usepackage{tabularx}
\usepackage{url}
\usepackage[altpo]{backnaur}

\usepackage{epsfig}
\usepackage{graphicx}
\usepackage{amsmath}
\usepackage{amssymb}
\usepackage{gensymb}
\usepackage{footmisc}
\usepackage{pifont}
\PassOptionsToPackage{table}{xcolor}
\usepackage{array}
\usepackage{url}

\usepackage{collcell}
\usepackage{xcolor}
\usepackage{colortbl}

\usepackage{listings}
\usepackage{xcolor}
\usepackage{fancyvrb}

\definecolor{Gray}{gray}{0.94} 

\newcommand*{\MinNumber}{-0.01}%

\newcommand{\AG}[1]{%

        \ifdim #1 pt > \MinNumber pt
            \textcolor{black}{#1}

        \fi

}

\ifCLASSOPTIONcompsoc
  \usepackage[nocompress]{cite}
\else
  \usepackage{cite}
\fi

\ifCLASSINFOpdf
\else
\fi

\begin{document}

\title{Lightweight Cross-Spectral Face Recognition via Contrastive Alignment and Distillation}
\author{Anjith~George,~\IEEEmembership{Member,~IEEE,}
        S\'ebastien~Marcel,~\IEEEmembership{Fellow,~IEEE}
\IEEEcompsocitemizethanks{\IEEEcompsocthanksitem All authors are with Idiap Research Institute, Martigny, Switzerland. S\'{e}bastien Marcel is also affiliated with Universit\'{e} de Lausanne (UNIL), Lausanne, Switzerland.
E-mail: \{anjith.george, sebastien.marcel\}@idiap.ch 
}
\thanks{Manuscript received April 19, 2005; revised August 26, 2015.}}

\markboth{Journal of \LaTeX\ Class Files,~Vol.~14, No.~8, August~2015}%
{Shell \MakeLowercase{\textit{et al.}}: Bare Demo of IEEEtran.cls for Biometrics Council Journals}

\IEEEtitleabstractindextext{%
\begin{abstract}

Heterogeneous Face Recognition (HFR) aims at matching face images captured across different sensing modalities, such as thermal-to-visible or near-infrared-to-visible, enhancing the usability of face recognition systems in challenging real-world conditions. Although recent HFR methods have achieved significant improvements in performance, many rely on computationally expensive models, making them impractical for deployment on resource-limited edge devices. In this work, we introduce a lightweight yet effective HFR framework by adapting a hybrid CNN-Transformer model originally developed for RGB homogeneous face recognition. Our approach enables efficient end-to-end training with only a small amount of paired heterogeneous data, while still maintaining strong performance on standard RGB face recognition benchmarks. This makes it suitable for both homogeneous and heterogeneous settings. Comprehensive experiments on several challenging HFR and face recognition benchmarks show that our method achieves state-of-the-art or competitive performance while keeping computational requirements low.
\end{abstract}

}

\maketitle

\IEEEdisplaynontitleabstractindextext

%
\IEEEpeerreviewmaketitle

\IEEEraisesectionheading{\section{Introduction}\label{sec:introduction}}

\IEEEPARstart{F}ace recognition (FR) has become a ubiquitous modality in biometric authentication, particularly in access control, due to its efficiency and non-intrusive nature. With advances in deep learning, especially convolutional neural networks (CNNs), face recognition has achieved near-human performance even in unconstrained settings~\cite{learned2016labeled}. However, most of the existing FR methods are designed for homogeneous environments, where both gallery and probe images are captured using visible-spectrum cameras.

In many practical use cases, such as surveillance, mobile authentication, or defense applications, relying only on visible-light imagery is insufficient. Images captured outside the visible spectrum, such as near-infrared (NIR) or thermal imagery, provide several advantages. For example, NIR is more resistant to lighting variations and more robust against spoofing attempts~\cite{li2007illumination,george2022comprehensive}. Despite these advantages, developing effective FR models for these modalities is difficult, mainly due to the lack of large-scale annotated heterogeneous paired datasets. Heterogeneous Face Recognition (HFR) addresses this challenge by enabling face matching across modalities, for instance, comparing thermal or NIR images with visible-light references~\cite{klare2012heterogeneous,anghelone2025beyond}. A key aspect of HFR is Cross-Spectral Face Recognition (CFR), which focuses on handling the appearance variations caused by spectral differences between imaging domains. CFR becomes especially important in low-light or long-range scenarios where visible imaging is unreliable.

Although deep neural networks (DNNs) have greatly improved Heterogeneous Face Recognition (HFR), the task remains challenging due to the large modality gap between source and target domains. Models trained on RGB images often fail to generalize to non-RGB inputs~\cite{he2018wasserstein}. Moreover, the collection of large-scale paired cross-modal datasets is both costly and difficult, which requires the development of methods capable of generalizing from limited training data. Many state-of-the-art HFR systems also rely on computationally heavy architectures, which are impractical for deployment on edge or mobile devices. Consequently, there is growing interest in lightweight models that maintain competitive accuracy while reducing computational overhead. Vision Transformers (ViTs) have demonstrated a strong capability to capture global dependencies~\cite{khan2022transformers}, making them a useful complement to CNN-based architectures. These architectures provide an opportunity to design compact yet effective HFR frameworks suitable for real-world, resource-limited environments.

In this work, we propose a parameter-efficient adaptation framework that extends pretrained RGB face recognition models to heterogeneous face recognition without increasing inference complexity. Instead of introducing modality-specific branches or additional network modules, our approach selectively adapts LayerNorm parameters and early convolutional layers while keeping the remainder of the backbone frozen. Combined with contrastive alignment and self-distillation, this strategy enables effective cross-spectral adaptation using only a small amount of paired heterogeneous data while preserving the original RGB recognition performance. We demonstrate this approach using the lightweight EdgeFace\cite{george2024edgeface} architecture as a backbone.

\noindent The main contributions of this work are as follows.

\begin{itemize}
\item We propose a parameter-efficient adaptation framework that extends pretrained RGB face recognition models to heterogeneous face recognition without increasing inference complexity.

\item We show that adapting only LayerNorm parameters and shallow layers, combined with contrastive alignment and self-distillation, enables effective cross-modal learning using limited paired data.
\item Extensive experiments across six heterogeneous benchmarks demonstrate competitive or state-of-the-art performance while maintaining significantly lower computational cost. Code is available at \footnote{\url{https://www.idiap.ch/paper/lightweighthfr}}.
\end{itemize}

\section{Related Work}
\textbf{Heterogeneous Face Recognition}: focuses on matching faces captured across different imaging modalities such as visible light (VIS), near-infrared (NIR), thermal cameras, or even hand-drawn sketches. The main challenge in HFR is the modality gap, meaning the substantial distribution shift between these modalities. This gap causes standard face recognition models trained solely on RGB images to perform poorly when applied to other domains. To address this issue, recent studies introduce a variety of solutions that generally fall into three categories: learning modality-invariant features, projecting data into a shared representation space, and generating source domain samples through synthesis-based methods.

Invariant feature based approaches focus on learning facial representations that remain stable across different modalities. Early studies focused on handcrafted descriptors such as Difference of Gaussian (DoG) filters, multi-scale LBP \cite{liao2009heterogeneous}, SIFT, and MLBP \cite{klare2010matching} to capture local texture cues. With the rise of deep learning, CNN-based models were introduced to learn modality-invariant features \cite{he2017learning,he2018wasserstein}, while other works proposed improved handcrafted features like the Local Maximum Quotient (LMQ) descriptor \cite{roy2018novel} or explored composite feature fusion at the score level \cite{liu2018composite}. In \cite{de2025towards}, the authors proposed a NIR-VIS heterogeneous face recognition framework that augments lightweight face-recognition models with Gabor filter–derived invariant features by appending the filter’s imaginary response as an additional input channel, followed by PCA-based dimensionality reduction and Mahalanobis-distance matching. This strategy improves NIR-VIS recognition performance across benchmark datasets while largely preserving VIS-domain performance and incurring only minimal computational overhead.

Common-space projection methods address the domain gap by mapping features from different modalities into a unified latent space. Classical techniques include Canonical Correlation Analysis (CCA) \cite{yi2007face}, Partial Least Squares (PLS) \cite{sharma2011bypassing}, and various forms of coupled regression \cite{lei2009coupled}, all of which use linear or nonlinear transformations to preserve discriminative properties while reducing the domain gap. More recent work leverages deep architectures with domain-specific units \cite{de2018heterogeneous}, domain-invariant modules \cite{george2024heterogeneous}, coupled attribute-aware loss functions \cite{liu2020coupled}, and semi-supervised collaborative representations \cite{liu2023modality}, enhancing alignment even under limited annotations or unpaired data. Further progress \cite{george2024modalities,george2023bridging} shows that conditioning intermediate feature maps can effectively bridge the modality gap, later extended toward modality-agnostic learning \cite{george2024modality}.

Synthesis-based methods adopt a different strategy by generating cross-modal images, typically translating inputs into the visible domain, so that standard face recognition models can be directly applied. Early solutions performed patch-level reconstruction via Markov Random Fields \cite{wang2008face} or employed manifold learning approaches such as LLE \cite{liu2005nonlinear}. The advent of GAN-based frameworks, including CycleGAN \cite{zhuUnpairedImagetoImageTranslation2017}, greatly improved this approach, enabling unpaired translation and photorealistic facial synthesis \cite{zhang2017generative,fu2021dvg}. More recent advances explore latent disentanglement \cite{liu2021heterogeneous}, memory-enhanced transformers for unsupervised reference-guided generation \cite{luo2022memory}, and plug-and-play modules like Prepended Domain Transformers (PDT) \cite{george2022prepended}, which improve cross-domain representation alignment without explicit image generation. Nonetheless, synthesis-based approaches often introduce significant computational overhead because they require both image translation module and a separate face recognition model for matching.

Recent advances in NIR-VIS heterogeneous face recognition have increasingly focused on reducing  the dependence on manually annotated labels, as acquiring large-scale labeled cross-domain datasets is both costly and impractical for real-world deployment. To address this, many recent works adopt unsupervised or semi-supervised learning strategies that exploit pseudo-label generation, contrastive learning, and prototype-based representation learning. For instance, \cite{yang2023robust} formulates the task as an unsupervised domain adaptation problem and introduces the RPC network, which combines NIR cluster-based pseudo-label sharing with both domain-specific and inter-domain contrastive learning to produce compact and domain-invariant representations, achieving over 99\% pseudo-label assignment accuracy and strong benchmark performance. Similarly, \cite{hu2024pseudo} addresses the problem in a semi-supervised setting through the LPL framework, which combines cross-domain pseudo-label association, intra-domain compact representation learning, and prototype-based inter-domain invariant learning to iteratively refine cluster structure and extract robust cross-domain identity features, reaching performance comparable to recent supervised methods. Along the same line, \cite{yang2023unsupervised} proposes HERE (HEterogeneous learning and Residual-invariant Enhancement), an unsupervised framework that employs a homogeneous-to-heterogeneous learning strategy, combining modality-adversarial contrastive learning, cross-modal pseudo-label estimation, refined contrastive learning, and residual-invariant feature enhancement to learn robust modality-invariant representations. Together, these studies demonstrate that competitive NIR-VIS recognition performance can be achieved with limited or no explicit identity supervision.

\textbf{Lightweight Face Recognition}: As mobile devices and edge computing platforms have become ubiquitous, face recognition (FR) research has increasingly prioritized compact models that provide strong performance under tight computational and memory constraints. This shift has driven the development of a wide range of efficient architectures adapted for FR. MobileFaceNets~\cite{chen2018mobilefacenets}, built on the MobileNet family~\cite{howard2017mobilenets,sandler2018mobilenetv2}, were among the first to achieve high performance with under 1M parameters. MixFaceNets~\cite{boutros2021mixfacenets} further improved efficiency by integrating MixConv~\cite{tan2019mixconv}, while ShiftFaceNet~\cite{wu2018shift} leveraged ShiftNet operations to achieve competitive results with just 0.78M parameters. ShuffleFaceNet~\cite{martindez2019shufflefacenet}, inspired by ShuffleNetV2~\cite{ma2018shufflenet}, introduced model variants ranging from 0.5M to 4.5M parameters without sacrificing accuracy. Neural architecture search has also advanced lightweight FR: PocketNet~\cite{boutros2022pocketnet}, designed via DARTS on CASIA-WebFace~\cite{yi2014learning} with multi-stage knowledge distillation (KD), and VarGFaceNet~\cite{yan2019vargfacenet}, winner of the ICCV 2019 LFR challenge~\cite{deng2019lightweight} used variable group convolutions to optimize efficiency. Recent work, SynthDistill~\cite{shahreza2024knowledge}, demonstrated that synthetic data \cite{george2025digi2real} combined with online KD can effectively train TinyFaR models~\cite{han2020model} to approximate high-performance teacher networks. GhostFaceNets~\cite{alansari2023ghostfacenets} reduced redundancy in convolutional operations, achieving extremely compact designs with as little as 61M FLOPs using depthwise convolutions. Most recently, EdgeFace~\cite{george2024edgeface} combined convolutional and transformer components leveraging EdgeNeXt~\cite{maaz2023edgenext} framework and employed low-rank linear layers to cut both parameters and FLOPs, achieving near–state-of-the-art results at a fraction of the complexity and securing the top position among compact models in the IJCB  EFaR 2023 challenge~\cite{kolf2023efar}.

\textbf{Lightweight Heterogeneous Face Recognition}: While lightweight FR architectures are ideal for edge applications, their adaptation to the more challenging Heterogeneous Face Recognition (HFR) has not been addressed in literature. Many current HFR systems rely on heavy backbones or synthesis-based pipelines, both of which introduce substantial computational costs that hinder deployment in resource-limited environments. To address this gap, we introduce a compact and efficient HFR framework tailored for edge devices,  with strong cross-modal performance while keeping computational demands and data requirements minimal.

\section{Proposed Approach}

\begin{figure*}[t!]
  \centering
      \includegraphics[width=0.95\linewidth]{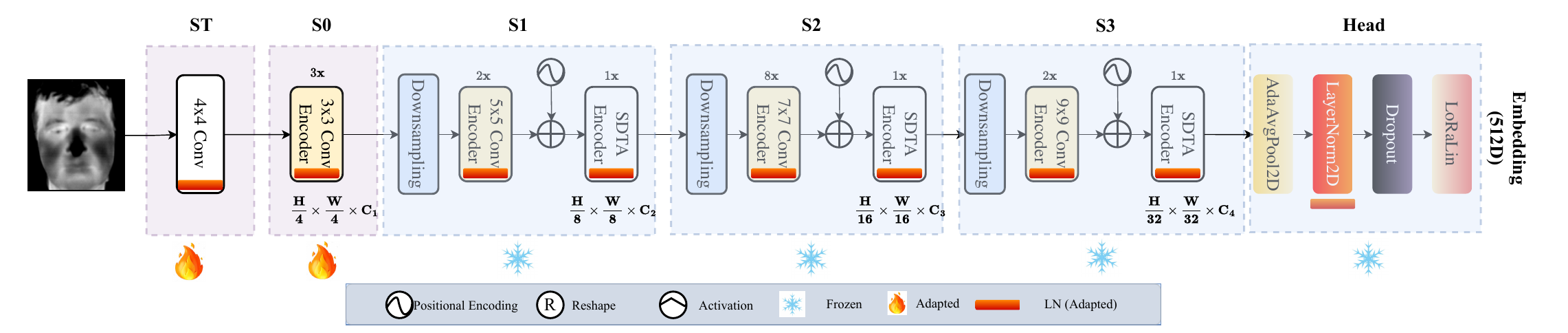}
  \caption{Model architecture of xEdgeFace models: The highlighted modules (LN-LayerNorm, ST-Conv. Stem, Stages-S0, S1, S2) are adapted while other network components remain frozen. The two loss components ensure modality alignment while preserving source-domain FR performance. Computational complexity remains unchanged in new models.}
  \label{fig:framework}
\end{figure*}

Heterogeneous face recognition (HFR) poses a significant challenge largely due to the limited availability of paired cross-modal training data. To address this, a widely used approach is to start with large-scale models pretrained on visible-spectrum (RGB) images and then adapt them to heterogeneous domains. However, directly fine-tuning these models on small HFR datasets often results in overfitting and catastrophic forgetting, where the model’s original RGB recognition performance deteriorates heavily. To mitigate this, prior works \cite{de2018heterogeneous,george2022prepended} have introduced architectural changes such as modality-specific branches or asymmetric processing pathways. While these designs help retain RGB performance, they also increase model size and introduce parameter redundancy, an undesirable trade-off when aiming for lightweight models. Further, many existing approaches assume a fixed representation for the RGB modality and force the other modality (e.g., NIR, thermal, sketch) to align with the source modality in the original latent space. This rigid alignment can limit performance, especially when cross-modal differences are highly nonlinear.

In this work, our goal is to develop a unified model that effectively handles both standard (homogeneous) and heterogeneous face recognition without adding noticeable computational overhead or reducing accuracy. We introduce a lightweight yet robust adaptation strategy for pretrained FR models that prevents catastrophic forgetting while enabling strong cross-modal matching performance. Unlike prior modulation-based approaches such as PDT, CAIM, and SSMB, which introduce modality-specific branches or additional inference-time modules, our method performs parameter-efficient adaptation within the existing pretrained backbone. Specifically, we selectively update LayerNorm parameters and early convolutional layers while keeping the remaining network frozen. This design avoids additional architectural complexity and maintains the original inference cost. Furthermore, the proposed self-distillation objective preserves the original RGB performance, reducing catastrophic forgetting during heterogeneous adaptation.

Our method builds upon EdgeFace \cite{george2024edgeface}, a lightweight architecture that integrates convolutional layers with transformer modules. The main idea in our approach is that Layer Normalization \cite{ba2016layer} plays a useful role in modality adaptation \cite{gatys2016image}. Instead of modifying the backbone structure or adding redundant pathways, we treat LayerNorm as a modulation mechanism that adjusts modality-specific statistics, enabling the network to learn discriminative features for both RGB and other modalities within a single shared architecture.

To accomplish our goal, we adopt a contrastive self-distillation training approach. The objective consists of two key components:

\begin{itemize}
    \item  Contrastive Modality Alignment, which encourages paired samples (e.g., RGB–NIR) to produce closer embeddings in the shared latent space promoting modality-invariant representations; and
    \item Self-Distillation Regularization, which preserves the pretrained model’s RGB accuracy by transferring its knowledge to the adapted model reducing catastrophic forgetting.
\end{itemize}

Together, these components allow effective fine-tuning on limited HFR data while retaining strong RGB performance and maintaining a lightweight design. The following subsections provide detailed explanations of the training objectives, backbone configuration, and implementation specifics.

\textbf{LayerNorm Adaptation}: Prior work has shown that the statistical characteristics of feature maps in deep neural networks (DNNs) encode key stylistic cues of images first demonstrated by Gatys et al.~\cite{gatys2016image}. This insight has made normalization layers crucial for stabilizing and improving deep model training. Layer Normalization (LayerNorm), introduced by Ba et al.~\cite{ba2016layer}, overcomes the limitations of batch normalization by computing normalization statistics over the feature dimension of each individual sample rather than across the batch. As a result, LayerNorm maintains consistent behavior during both training and inference, making it well-suited for settings involving variable input structures or non-i.i.d. data. In large language models (LLMs) and their multimodal variants (MLLMs), recent work by Zhao et al.~\cite{zhao2024tuning} has shown that selectively fine-tuning LayerNorm parameters inside attention blocks can yield significant efficiency gains while reducing computational cost. This approach outperforms several parameter-efficient fine-tuning techniques, such as Low-Rank Adaptation (LoRA) \cite{hu2022lora}. LayerNorm’s scale and shift parameters serve as natural modulation points: small updates to these parameters can reshape the distribution of intermediate features across layers without modifying the pretrained backbone. As observed in Xu et al.~\cite{xu2019understanding} and Zhao et al.~\cite{zhao2024tuning}, the expected gradient of LayerNorm diminishes with network depth, while the gradient variance remains low, properties indicative of good generalization. These low-magnitude, low-variance gradients make it possible to adapt the model to new domains by updating only the LayerNorm parameters, preserving the stability of the pretrained network. This characteristic is especially valuable in our setting, where robust cross-modal adaptation must be achieved without extensively altering the underlying model weights, preventing catastrophic forgetting.

For modalities such as NIR or Thermal, the modality gap is primarily spectral/photometric rather than geometric. The facial structure remains largely unchanged, while the image statistics seen by the network such as intensity distribution, local contrast, and channel-wise feature responses shift substantially across modalities. In pretrained face recognition models, these low-level statistics are mainly processed in the early convolutional layers and normalization layers. Adapting these components therefore provides a principled way to compensate for modality-dependent shifts at the feature-statistics level, while leaving most of the deeper identity-discriminative representation unchanged. 

\textbf{Problem Setting}: We start with a pretrained face recognition network $F$ , whose parameters $\Theta_{\text{FR}}$ have been learned from a large-scale RGB (visible-spectrum) dataset.
 Let $(X_{s_i}, X_{t_i}, y_i)$ denote a triplet consisting of a pair of images $X_{s_i}$ and $X_{t_i}$ from the source (e.g., RGB) and target (e.g., NIR or thermal) modalities respectively, and a binary identity label $y_i \in \{0, 1\}$, where $y_i = 1$ indicates that both images correspond to the same identity and $y_i = 0$ otherwise.

The goal is to adapt $F$ into a heterogeneous face recognition network $\hat{F}$, parameterized by $\Theta_{\text{HFR}}$, such that the resulting embeddings $e_{s_i} = \hat{F}(X_{s_i})$ and $e_{t_i} = \hat{F}(X_{t_i})$ are well-aligned in a shared embedding space if they belong to the same identity, while also preserving the discriminative ability of the original model $F$ on the source modality.

We initialize $\hat{F}$ with the pretrained parameters $\Theta_{\text{FR}}$, and decompose $\Theta_{\text{HFR}}$ into three disjoint subsets:
\begin{equation}
\Theta_{\text{HFR}} = \left\{ \Theta_{\text{LN}}^{(1:K)}, \Theta_{\text{Adapted}}, \Theta_{\text{Frozen}} \right\},
\end{equation}
where $\Theta_{\text{LN}}^{(1:K)}$ denotes the set of all LayerNorm parameters (from $K$ layers), $\Theta_{\text{Adapted}}$ includes all trainable parameters except LayerNorms, and $\Theta_{\text{Frozen}}$ refers to the set of parameters that remain fixed during training.

To enforce alignment between embeddings from different modalities, we use a cosine-based contrastive loss defined as follows:
\begin{equation}
\label{eq:contrastive}
\begin{aligned}
\mathcal{L}_{\text{C}}(e_{s_i}, e_{t_i}, y_i) =\; & y_i \cdot \left(1 - \cos(e_{s_i}, e_{t_i})\right) \\
& + (1 - y_i) \cdot \max\left(0, \cos(e_{s_i}, e_{t_i}) - m\right),
\end{aligned}
\end{equation}
where $\cos(e_{s_i}, e_{t_i}) = \frac{e_{s_i} \cdot e_{t_i}}{\|e_{s_i}\|_2 \|e_{t_i}\|_2}$ is the cosine similarity between the embeddings and $m \in [0, 1]$ is a contrastive margin.

To retain the model’s original recognition ability on the source modality, we incorporate a self-distillation loss that guides the adapted model $\hat{F}$ to remain consistent with the pretrained model $F$ on source-domain embeddings. Formally, this loss is defined as:

\begin{equation}
\label{eq:self_distillation}
\mathcal{L}_{\text{SDL}}(e_{F_{s_i}}, e_{\hat{F}_{s_i}}) = 1 - \cos(e_{F_{s_i}}, e_{\hat{F}_{s_i}}),
\end{equation}
where $e_{F_{s_i}} = F(X_{s_i})$ is the frozen embedding from the original model and $e_{\hat{F}_{s_i}} = \hat{F}(X_{s_i})$ is the adapted embedding for the same image.

The overall training objective for the adapted network ($\hat{F}$) combines the contrastive loss for modality alignment with the self-distillation loss that preserves source-domain performance, resulting in the following formulation:

\begin{equation}
\label{eq:total_loss}
\begin{aligned}
\mathcal{L}_{\text{total}} = &\; (1 - \lambda) \cdot \mathcal{L}_{\text{C}}(e_{s_i}, e_{t_i}, y_i) \\
& + \lambda \cdot \mathcal{L}_{\text{SDL}}(e_{F_{s_i}}, e_{\hat{F}_{s_i}}),
\end{aligned}
\end{equation}
where $\lambda \in [0, 1]$ is a balancing hyperparameter controlling the trade-off between cross-modal alignment and self-regularization.

In all of our experiments, we set $\lambda = 0.75$ and the margin $m = 0$ unless otherwise indicated. This configuration empirically provided the best balance between adapting to the heterogeneous domain and retaining source modality performance.

Although our implementation uses EdgeFace as the backbone, the proposed contrastive alignment and self-distillation objectives are architecture-agnostic. In principle, they can be applied to other pretrained face recognition networks, enabling similar parameter-efficient heterogeneous adaptation without modifying the underlying architecture.

\textbf{Face Recognition Backbone} We use the pretrained EdgeFace \cite{george2024edgeface} model as our face recognition (FR) backbone. EdgeFace is a hybrid convolutional–transformer architecture that employs LayerNorm instead of the more common BatchNorm, enabling stable training across modalities. The model is trained on the large-scale WebFace12M dataset \cite{zhu2021webface260m}, which includes over 12 million RGB images from more than 600,000 identities. All input faces are resized to $112 \times 112$ and aligned via a similarity transform to standardize eye locations. For thermal images, which are single-channel, we replicate the channel three times to match the RGB input format required by the backbone.

\textbf{Implementation Details} Our HFR framework incorporates a frozen copy of the pretrained EdgeFace model as a regularization teacher network, guiding the fine-tuning of a shallow, trainable surrogate network through self-distillation (Figure \ref{fig:framework}). The surrogate is initialized with the pretrained weights, and only selected early modules including LayerNorm layers, are unfrozen for adaptation, while the remaining components stay frozen. To further improve cross-modal consistency, we apply a contrastive loss on the surrogate’s embeddings for both RGB and thermal inputs.

The framework is implemented in PyTorch and built upon the Bob library \cite{bob2012, bob2017}\footnote{\url{https://www.idiap.ch/software/bob/}}. Training uses the Adam optimizer with a learning rate of $1 \times 10^{-4}$, a batch size of 256, and runs for 20 epochs. The contrastive loss margin $m$ is set to 0, and the weighting factor $\lambda$ is fixed at 0.75 for all experiments. Although both pretrained and surrogate networks are used during training, only the adapted surrogate model is required for inference.

\section{Experiments}

This section reports the results of an extensive set of experiments conducted using the proposed framework. We evaluate heterogeneous face recognition (HFR) performance on established benchmarks and compare our method against state-of-the-art approaches. To verify that the adaptation process does not introduce catastrophic forgetting, we also assess performance on standard face recognition datasets. Across all experiments, cosine distance is used as the similarity metric for evaluation.

\subsection{Datasets and Protocols}

For our evaluations, we used the following datasets:

\textbf{Tufts Face Dataset:} The Tufts Face Database~\cite{panetta2018comprehensive} contains a broad collection of facial images captured across multiple modalities, making it well-suited for heterogeneous face recognition tasks. In our experiments, we followed the \textit{VIS-Thermal} protocol and used the thermal subset of the dataset. Tufts includes 113 identities (39 males and 74 females) covering diverse demographic groups, with each subject represented across several modalities. Following the protocol in~\cite{fu2021dvg}, we randomly selected 50 identities for training (with 45 for training and 5 as validation) and used the remaining subjects for testing.

\textbf{MCXFace Dataset:} The MCXFace dataset~\cite{george2022prepended,mostaani2020high} consists of facial images from 51 participants, collected under different illumination conditions across three sessions and multiple sensing channels. These channels include RGB, thermal, near-infrared (850 nm), short-wave infrared (1300 nm), depth, and depth estimated from RGB. The dataset provides five folds, each created by randomly splitting identities into training and development sets. Our evaluations focus on the challenging \textit{VIS-Thermal} protocols, which serve as standard benchmarks for cross-modal face recognition on this dataset.

\textbf{Polathermal Dataset:} The Polathermal dataset~\cite{hu2016polarimetric}, collected by the U.S. Army Research Laboratory (ARL), is an HFR dataset containing both polarimetric long-wave infrared (LWIR) imagery and visible-spectrum color images for 60 subjects. In addition to polarimetric data, the dataset provides conventional thermal images for each subject. Following the five-fold partitioning protocol introduced in~\cite{de2018heterogeneous}, we use the conventional thermal images, assigning 25 identities for training and the remaining 35 identities for testing.

\textbf{SCFace Dataset:} The SCFace dataset~\cite{grgic2011scface} includes high-quality enrollment images paired with low-quality probe images captured in realistic surveillance environments using multiple cameras. It is organized into four evaluation protocols: \emph{close}, \emph{medium}, \emph{combined}, and \emph{far}, with the ``far'' protocol presenting the highest level of difficulty. In total, the dataset contains 4,160 static images from 130 subjects, recorded in both visible and infrared domains.

\textbf{CUFSF Dataset:} The CUHK Face Sketch FERET Database (CUFSF)~\cite{zhang2011coupled} consists of 1,194 facial photographs from the FERET dataset~\cite{phillips1998feret}, each paired with an artist-drawn sketch. Due to the stylized and exaggerated nature of the sketches, CUFSF poses a challenging HFR setting. Following the protocol in~\cite{fang2020identity}, we use 250 identities for training and evaluate on the remaining 944 identities.

\textbf{CASIA NIR-VIS 2.0 Dataset:} The CASIA NIR-VIS 2.0 Face Database~\cite{li2013casia} contains images from 725 subjects captured in both visible (VIS) and near-infrared (NIR) modalities. Each subject has around 1-22 VIS images and 5-50 NIR images. Experiments follow a fixed 10-fold cross-validation protocol with 360 identities used for training. The gallery and probe sets include 358 distinct individuals, ensuring no identity overlap between training and evaluation.

\textbf{Metrics:} We evaluate model performance using several standard metrics commonly adopted in the heterogeneous face recognition literature. These include the Area Under the Curve (AUC), Equal Error Rate (EER), Rank-1 identification accuracy, and Verification Rates at multiple false acceptance rates (0.01\%, 0.1\%, 1\%, and 5\%). For datasets with multiple folds, we report the mean performance along with the corresponding standard deviation across folds.

\subsection{Model Complexity} The main objective of this work is the development of lightweight models for heterogeneous face recognition. Therefore, it is important to compare the computational footprint of our proposed models with those commonly used in prior literature. We assess computational efficiency using two standard metrics: the number of floating-point operations (in terms of GFLOPs) and the total number of parameters (in millions, denoted as MPARAMs). As summarized in Table~\ref{tab:computational_complexity}, the xEdgeFace \cite{xedgeface} variants achieve significantly lower computational cost and parameter count, highlighting their suitability for deployment in resource-limited environments. Throughout the remaining comparisons, it is important to note that the xEdgeFace base model has \textit{one-third the parameters} and demands \textit{roughly one-twentieth of the compute} of the state-of-the-art models.

\begin{table}[!htb]
  \caption{Comparison of computational complexity between the proposed method and state-of-the-art HFR approaches, reported in terms of floating point operations (GFLOPs) and number of parameters (MPARAMs). }
  \label{tab:computational_complexity}
  \centering
  \resizebox{0.7\columnwidth}{!}{%
\begin{tabular}{lcc}
\toprule
{} &  \textbf{GFLOPS} &  \textbf{MPARAMS} \\
\midrule
CAIM(1-3) \cite{george2023bridging} &    26.3 &    65.6 \\
DIU \cite{george2024heterogeneous} & 24.2 & 65.2 \\
SSMB \cite{george2024modality} &24.2 &65.5 \\
PDT \cite{george2022prepended} &24.2 &65.2 \\ \midrule
\rowcolor{Gray}
\textbf{xEdgeFace - Base} & 1.39 & 18.23 \\
\rowcolor{Gray}
\textbf{xEdgeFace - S} ($\gamma = 0.5$) & 0.31 & 3.65 \\
\rowcolor{Gray}
\textbf{xEdgeFace - XS} ($\gamma = 0.6$)  & 0.15 & 1.77 \\
\rowcolor{Gray}
\textbf{xEdgeFace - XXS} & 0.09 & 1.24 \\
\bottomrule
\end{tabular}
}
\end{table}
\subsection{\textbf{Ablation Studies}}

Given the large set of design choices and hyperparameters involved, we begin with a comprehensive ablation study to examine the impact of each component on overall model performance. All ablation experiments are conducted on the Tufts Face Dataset following the VIS-Thermal protocol, which represents one of the most challenging heterogeneous face recognition (HFR) settings due to its large modality gap. In the experiments that follow, adapted HFR versions of the base models are denoted as \emph{xEdgeFace}.

\textbf{Adapting Different Sets of Layers:} To identify the most effective subset of layers for adaptation, we perform a series of controlled ablation experiments by selectively unfreezing the LayerNorm (LN) layers, the initial convolutional stem (ST), and successive backbone stages: Stage~0 (S0), Stage~1 (S1), and Stage~2 (S2). In addition to cumulative configurations, we also evaluate individual stages and deeper-stage adaptation settings to better isolate the contribution of each component. The results, presented in Table~\ref{tab:ablation_layer_tufts}, show that adapting only the LayerNorm layers already provides a significant improvement over the pretrained baseline, while incorporating the stem and early stages further enhances performance. Among single-stage adaptations, S0 performs best (51.76\%), outperforming S1 and S2, and the same trend is observed when combined with LayerNorm, where (LN, S0) (54.92\%) clearly surpasses (LN, S1) and (LN, S2). The best overall result is obtained with (LN, ST, S0) (56.03\%), whereas adapting deeper stages provides limited additional benefit and slightly degrades performance. These results indicate that the main gains come from LayerNorm and early-stage adaptation, which offer the best balance between heterogeneous face recognition performance and parameter efficiency in training. At the same time, the average face recognition accuracy remains nearly unchanged across all configurations (96.6--96.8\%), confirming that the proposed adaptation improves heterogeneous verification without sacrificing standard RGB recognition.

\begin{table}[h]
  \centering
  \caption{Ablation study on the Tufts Face Dataset using different configurations of adapted layers. The verification rate on the Tufts Face dataset and the average accuracy on the face recognition benchmarks are shown.}
  \label{tab:ablation_layer_tufts}
  \resizebox{0.47\textwidth}{!}{

\begin{tabular}{lcc}
\toprule
\textbf{Adapted Layers} & \textbf{VR@FAR=0.01$\%$} & \textbf{Avg. FR Acc. (\%)} \\
\midrule
LN & 32.47 & 96.62 $\pm$ 0.93 \\
ST & 13.73 & 96.83 $\pm$ 0.90 \\
LN,ST & 48.79 & 96.64 $\pm$ 0.95 \\
\textbf{LN,ST,S0} & \textbf{56.03} & 96.72 $\pm$ 0.92 \\
LN,ST,S0,S1 & 53.62 & 96.71 $\pm$ 0.82 \\
LN,ST,S0,S1,S2 & 52.88 & 96.71 $\pm$ 0.85 \\ \midrule
S0 & 51.76 & 96.56 $\pm$ 0.88 \\
S1 & 48.05 & 96.78 $\pm$ 0.85 \\
S2 & 34.69 & 96.66 $\pm$ 0.90 \\ \midrule
LN,S0 & 54.92 & 96.74 $\pm$ 0.85 \\
LN,S1 & 44.53 & 96.75 $\pm$ 0.85 \\
LN,S2 & 40.45 & 96.65 $\pm$ 0.89 \\ \midrule
S1,S2 & 44.16 & 96.65 $\pm$ 0.87 \\
S0,S1,S2 & 49.17 & 96.68 $\pm$ 0.87 \\
ST,S0,S1,S2 & 44.90 & 96.69 $\pm$ 0.86 \\
LN,S0,S1,S2 & 53.99 & 96.69 $\pm$ 0.90 \\

\bottomrule
\end{tabular}
}

\end{table}

\textbf{Effect of Varying $\lambda$:} The hyperparameter $\lambda$ controls the balance between the supervision signal from the pretrained model and the modality alignment objective, both essential components for effective heterogeneous face recognition (HFR) task. Table~\ref{tab:ablation_lambda_tufts} summarizes the results of this ablation experiment. Setting $\lambda = 0$ emphasizes only the modality alignment term, removing all guidance from the pretrained network. Although this encourages alignment to the new modality, it quickly leads to overfitting due to the limited size of the training set. In contrast, using $\lambda = 1$ depends entirely on pretrained supervision, resulting in inadequate cross-modal alignment and poor HFR performance. The values of $\lambda = 0.50$ and $\lambda = 0.75$ both provide strong results; however, $\lambda = 0.75$ achieves the most favorable trade-off by placing slightly more weight on the pretrained guidance, which is important given the small fine-tuning dataset. This choice not only improves HFR accuracy but also reduces catastrophic forgetting, preserving performance on the original RGB domain ( Table~\ref{tab:tufts_fr_perf}).

\begin{table}[h]
  \centering
  \caption{Ablation study with varying values of hyperparameter $\lambda$.}
  \label{tab:ablation_lambda_tufts}
  \resizebox{0.35\textwidth}{!}{
\begin{tabular}{lcccc}
\toprule
$\lambda$ &    \textbf{AUC} &    \textbf{EER} &     \textbf{Rank-1} & \textbf{VR@FAR=1$\%$} \\
\midrule
0.00 &  94.52 &  12.23 &  52.96 &  46.57 \\
0.25 &  97.87 &   7.24 &  79.53 &  81.82 \\
0.50 &  \textbf{98.46} &   \textbf{5.95} &   \textbf{83.3 }&  84.79 \\
0.75 &  97.91 &   6.68 &  82.59 &  \textbf{86.83} \\
1.00 &  87.47 &  20.22 &  42.19 &  42.86 \\
\bottomrule
\end{tabular}}
\end{table}

\textbf{Effect of Training Set Size}: In Table ~\ref{tab:tufts_percentage}, we investigate how the number of subjects influences performance. In particular, our training protocol already operates in a highly data-efficient regime: using 100\% of the available training data corresponds to only 45 subjects with paired RGB–Thermal samples: orders of magnitude fewer identities than typical RGB face recognition models, which are often trained on more than 100K identities. Despite this limited supervision, our method achieves strong results, showing that the method can still learn useful representations from limited paired data.

\begin{table}[h]
\centering
\caption{Experimental results using different fractions of the training set on the Tufts Face Dataset.}
\label{tab:tufts_percentage}
\resizebox{0.99\columnwidth}{!}{

\begin{tabular}{lccccc}
\toprule
\textbf{\% of Training Data} & \textbf{Number of Subjects} & \textbf{AUC} & \textbf{EER} & \textbf{Rank-1} & \textbf{VR@FAR=1$\%$}  \\
\midrule
100\% & 45 & 97.91 &  6.68 & 82.59 & 86.83 \\
50\%  & 22 & 97.32 &  7.65 & 79.17 & 81.82 \\
20\%  &  9 & 95.71 & 11.13 & 70.38 & 74.40 \\
10\%  &  4 & 92.70 & 15.62 & 58.89 & 61.04 \\
5\%   &  2 & 90.15 & 18.37 & 48.11 & 46.94 \\
\bottomrule
\end{tabular}
}
  
\end{table}

\begin{figure}[t!]
  \centering
      \includegraphics[width=0.95\linewidth]{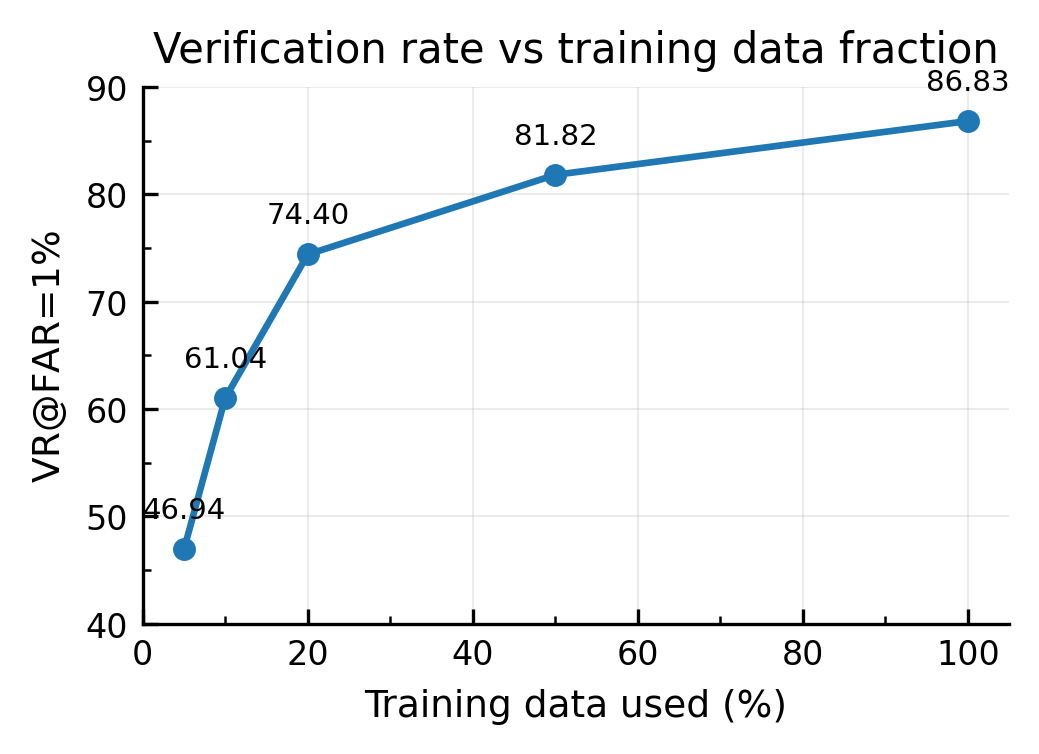}
  \caption{ Performance evolution using different fractions of the training data}
  \label{fig:tufra_trainfrac}
\end{figure}
To further evaluate the model, we progressively reduce the training set and show the performance trend in Fig.~\ref{fig:tufra_trainfrac}. As expected, performance decreases as less training data is used; however, the method remains relatively stable at moderate fractions and only shows a sharper decline below 20\% of the data (less than 10 subjects in the training set). This indicates that our approach is resilient under constrained data availability while also suggesting clear room for further improvement with more paired data.

\textbf{Experiments with Other EdgeFace Variants:} Although our primary experiments use the EdgeFace-Base model, we additionally evaluated the proposed adaptation strategy on smaller variants of the architecture. Table~\ref{tab:arch_comp} compares the performance of the original pretrained models with their adapted versions, denoted as \emph{xEdgeFace}. The results show that absolute heterogeneous face recognition (HFR) performance scales with the quality of the pretrained weights. Nonetheless, our adaptation method delivers substantial improvements across all model sizes, achieving relative gains of 103\%, 194\%, 257\%, and 361\% from the largest to the most compact variant. These findings demonstrate both the scalability and robustness of the proposed approach, showcasing its effectiveness even when applied to extremely lightweight model variants.

\begin{table}[h]
  \centering
  \caption{Comparison with different variants of EdgeFace}
  \label{tab:arch_comp}
  \resizebox{0.47\textwidth}{!}{
\begin{tabular}{lcccc}
\toprule
\textbf{Model} & \textbf{AUC} & \textbf{EER} & \textbf{Rank-1} & \textbf{VR@FAR=1\%} \\
\midrule
EdgeFace - Base & 87.44 & 20.37 & 42.73 & 42.86 \\
\rowcolor{Gray}
\textbf{xEdgeFace - Base} & 97.91 & 6.68 & 82.59 & \textbf{86.83} ($\uparrow\,\textbf{103}\%$) \\
EdgeFace - S ($\gamma = 0.5$) & 80.93 & 27.30 & 24.78 & 25.05 \\
\rowcolor{Gray}
\textbf{xEdgeFace - S} ($\gamma = 0.5$) & 96.93 & 8.89 & 71.10 & \textbf{73.65} ($\uparrow\,\textbf{194}\%$) \\
EdgeFace - XS ($\gamma = 0.6$) & 77.76 & 30.24 & 18.67 & 19.11 \\
\rowcolor{Gray}
\textbf{xEdgeFace - XS} ($\gamma = 0.6$) & 96.28 & 10.02 & 68.22 & \textbf{68.27} ($\uparrow\,\textbf{257}\%$) \\
EdgeFace - XXS & 75.72 & 31.73 & 17.41 & 12.80 \\
\rowcolor{Gray}
\textbf{xEdgeFace - XXS} & 95.14 & 11.35 & 60.50 & \textbf{59.00} ($\uparrow\,\textbf{361}\%$) \\
\bottomrule
\end{tabular}
}
\end{table}

\textbf{Face Recognition Performance of Adapted HFR Models:} To assess the face recognition (FR) capability of the adapted models beyond the heterogeneous setting, we evaluate the \textbf{xEdgeFace} variants on standard FR benchmarks. Specifically, we report accuracies on LFW~\cite{huang2008labeled}, CA-LFW~\cite{zheng2017cross}, CP-LFW~\cite{zheng2018cross}, CFP-FP~\cite{sengupta2016frontal}, and AgeDB-30~\cite{moschoglou2017agedb}. We compare xEdgeFace models adapted under both the VIS--NIR and VIS--Thermal HFR settings, with the latter representing the most challenging domain shift in our experiments. As shown in Table~\ref{tab:tufts_fr_perf}, the adapted model maintains FR performance that is nearly identical to the original EdgeFace backbone, even after adaptation to the severe VIS-Thermal scenario. At the same time, xEdgeFace achieves substantial improvements in HFR accuracy, demonstrating strong performance in both homogeneous and cross-modal recognition tasks. This behavior highlights the effectiveness of the self-distillation component, which acts as a regularizer that mitigates catastrophic forgetting and preserves the discriminative power of the pretrained RGB model. Overall, the proposed training strategy successfully extends the model’s capabilities to heterogeneous face recognition without degrading its original FR performance, thereby showcasing reliable performance in both homogeneous and heterogeneous settings.

\begin{table}[h]
  \centering
  \caption{Face recognition performance of the pretrained and adapted model.}
  \label{tab:tufts_fr_perf}
  \resizebox{0.49\textwidth}{!}{
\begin{tabular}{lccccc}
\toprule

                  \textbf{Model} &           \textbf{LFW} ~\cite{huang2008labeled} &         \textbf{CALFW}~\cite{zheng2017cross} &         \textbf{CPLFW}~\cite{zheng2018cross} &        \textbf{CFP-FP}~\cite{sengupta2016frontal} &      \textbf{AGEDB-30}~\cite{moschoglou2017agedb} \\
\midrule

EdgeFace - Base	 & 99.83 ± 0.24	 & 96.07 ± 1.03	 & 93.75 ± 1.16	& 97.01 ± 0.94	& 97.60 ± 0.70 \\
\textbf{xEdgeFace - Base} (VIS-Thermal) &  99.78 ± 0.27 &  95.85 ± 1.16 &  93.62 ± 1.31 &  96.83 ± 0.99 &  97.50 ± 0.88  \\
\textbf{xEdgeFace - Base} (VIS-NIR) & 99.82 ± 0.26 & 96.07 ± 0.99 & 93.78 ± 1.24 & 96.94 ± 0.97 & 97.28 ± 0.83 \\
\bottomrule
\end{tabular}
}
\vspace{-2mm}
\end{table}

\begin{figure}[t!]
  \centering
      \includegraphics[width=0.95\linewidth]{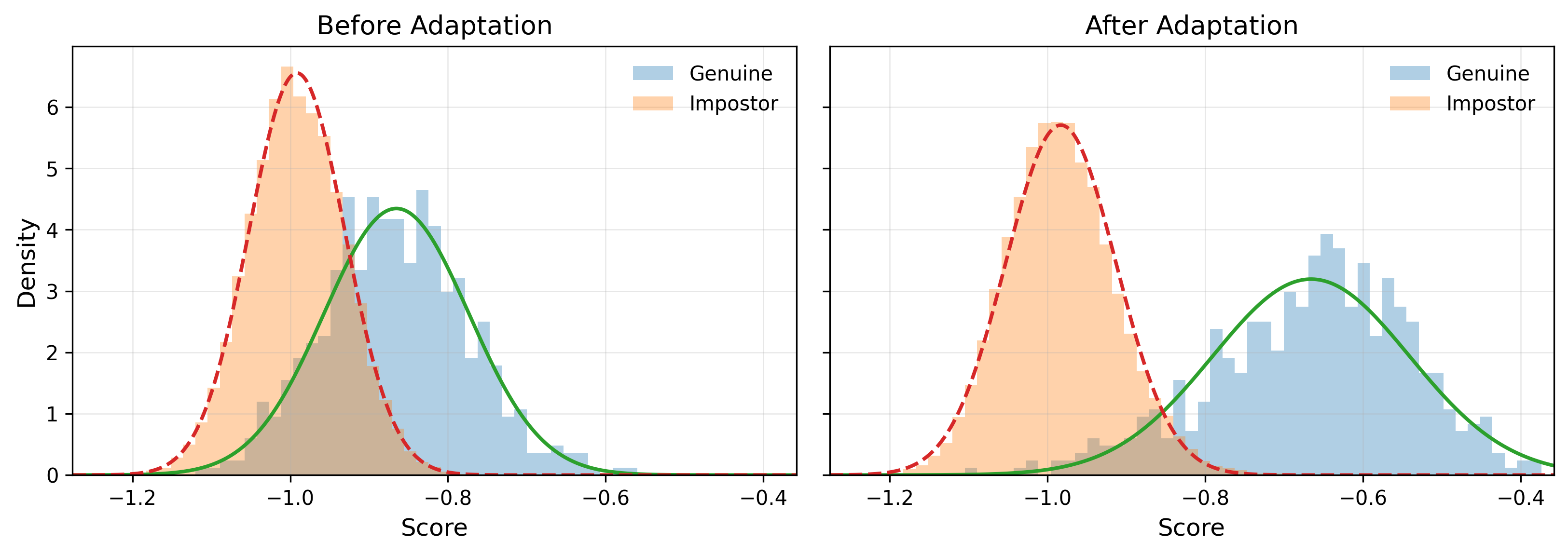}
  \caption{Genuine and Impostor Score Distribution Before and After Adaptation}
  \label{fig:score_before_after}
\end{figure}

\textbf{Visualizations} In this section, we compare the score distributions of the models before and after the adaptation process. Figure \ref{fig:score_before_after} shows the score distributions on the Tufts dataset under the VIS–Thermal protocol, illustrating the baseline performance where the genuine and impostor VIS–Thermal pairs are plotted. Prior to adaptation, the genuine score distribution heavily overlaps with the impostor distribution. After adaptation, however, the genuine distribution shifts to the right, indicating improved separability. This clearly demonstrates the improvement achieved by our proposed pipeline in the challenging VIS–Thermal matching scenario.

\begin{figure}[t!]
  \centering
      \includegraphics[width=0.95\linewidth]{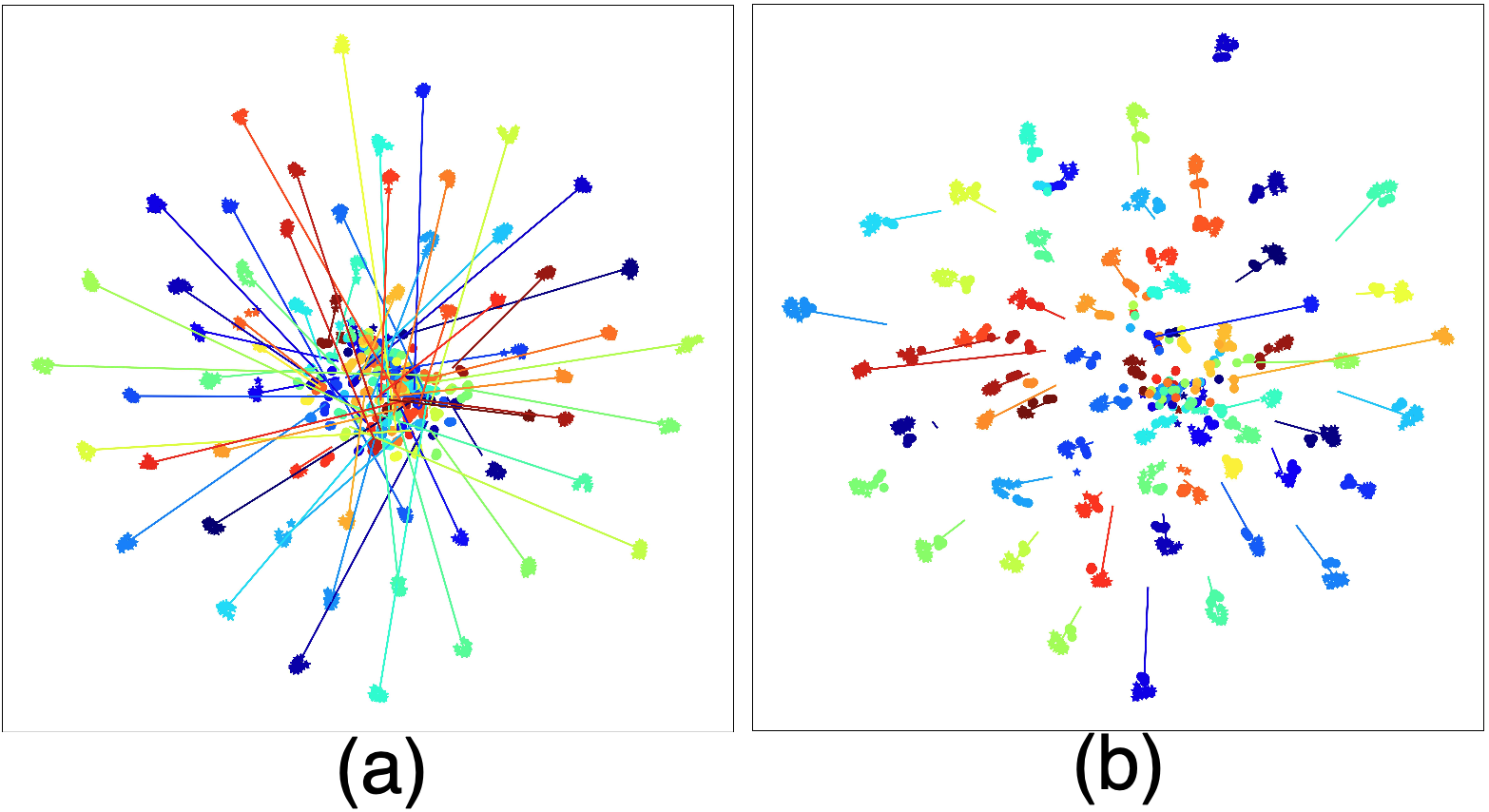}
  \caption{t-SNE plots of visible and thermal images at different stages of the pipeline, where each color represents a distinct identity. Lines connect the cluster centers of the visible and thermal images for each identity. (a) shows the embedding space before adaptation, while (b) shows the final embedding space. As observed, the identity clusters align much more closely in the final embedding space.}
  \label{fig:tsne_before_after}
\end{figure}

We further examine the t-SNE distribution of the embeddings before and after adaptation. The t-SNE plots in Fig. \ref{fig:tsne_before_after} illustrate how the embedding space evolves after the training process. After adaptation, the identities form tighter and more coherent clusters and align more closely with the source-domain embeddings.

\begin{figure}[t!]
  \centering
      \includegraphics[width=0.95\linewidth]{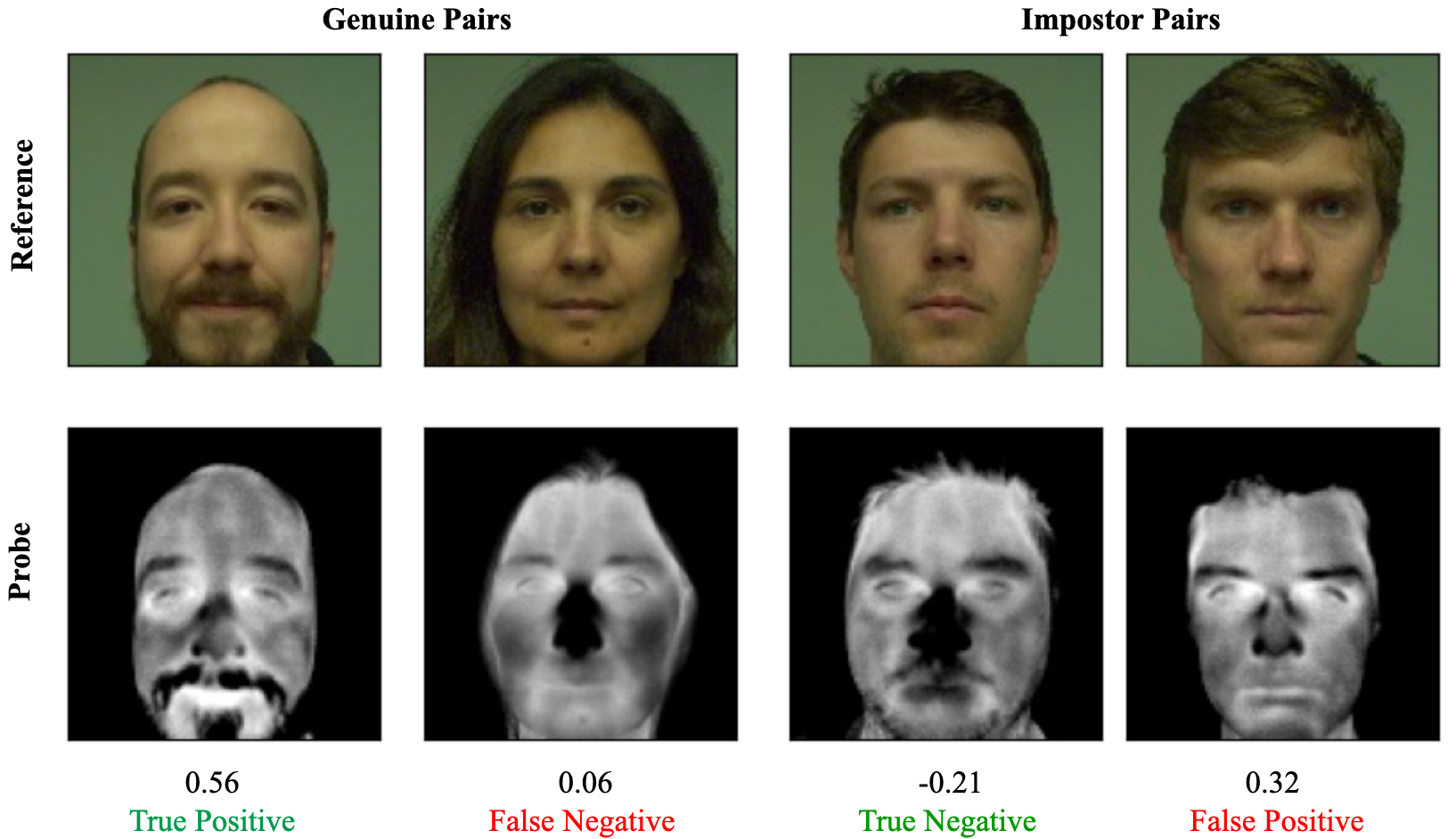}
  \caption{Illustration of successful and failure cases in the MCXFace VIS–Thermal face-matching scenario. The reported scores correspond to the cosine similarity computed between each reference–probe pair.}
  \label{fig:success_failure}
\end{figure}

We also present success and failure cases from the VIS–Thermal protocol, one of the most extreme and challenging recognition scenarios in the MCXFace dataset (Fig. \ref{fig:success_failure}). These examples are particularly useful for understanding the failure cases. As shown, the cold temperature in the nose region distorts key discriminative cues, leading to false negatives. In the false-positive examples, although the match scores remain relatively low, the similar facial structure, combined with the absence of rich textural information can still result in incorrect high match scores.

\subsection{Comparison with State-of-the-art}

In this section, we provide a comparative evaluation of the proposed \emph{xEdgeFace} model against state-of-the-art heterogeneous face recognition (HFR) approaches reported in the literature. However, it is important to remember that \emph{xEdgeFace} is substantially more lightweight than the competing models, highlighting the efficiency and practicality of our method. For all experiments, we use the xEdgeFace-Base variant with two adaptation configurations: \emph{(LN, ST)} and \emph{(LN, ST, S0)}. The adaptation loss weight $\lambda$ is fixed at 0.75 across all evaluations to maintain consistency, although this value can be further tuned for specific datasets depending on their size and the desired balance between adaptation strength and face recognition performance.

\textbf{Experiments with the Tufts Face Dataset:} Table~\ref{tab:tufts} reports the performance of \textbf{xEdgeFace} alongside state-of-the-art methods on the VIS-Thermal protocol of the Tufts Face Dataset. This dataset is particularly challenging due to substantial pose variations, especially extreme yaw angles that negatively impact both visible-spectrum and heterogeneous face recognition systems. Despite these challenges, the xEdgeFace model using the \emph{(LN, ST, S0)} configuration achieves the highest verification rate (69.02\%) and Rank-1 accuracy (82.59\%), while remaining extremely lightweight compared to competing approaches.

\begin{table}[h]
  \centering
  \caption{Experimental results on VIS-Thermal protocol of the Tufts Face dataset.}
  \label{tab:tufts}
  \resizebox{0.99\columnwidth}{!}{
  \begin{tabular}{lccc}
    \toprule
    \textbf{Method} & \textbf{Rank-1} & \textbf{VR@FAR=1$\%$} & \textbf{VR@FAR=0.1$\%$}  \\ \midrule
      LightCNN \cite{Wu2018ALC} & 29.4 & 23.0 & 5.3 \\
      DVG \cite{fu2019dual} & 56.1 & 44.3 & 17.1 \\
      DVG-Face \cite{fu2021dvg} & 75.7& 68.5 & 36.5 \\ 
      DSU-Iresnet100 \cite{george2022prepended} & 49.7 & 49.8 & 28.3 \\   
      PDT \cite{george2022prepended} & 65.71 & 69.39 & 45.45 \\ 
      CAIM \cite{george2023bridging}  & 73.07 & 76.81 & 46.94 \\ 
      SSMB \cite{george2024modality} (N=1) & 75.04 & 78.29 & 53.99 \\
      SSMB \cite{george2024modality} (N=2) & 78.46 & 80.33   &54.55 \\
      \midrule
      \rowcolor{Gray}
      \textbf{xEdgeFace-Base (LN, ST)}  & 75.76 & 79.59 & 62.89 \\
      \rowcolor{Gray}
      \textbf{xEdgeFace-Base (LN, ST, S0)}  &\textbf{82.59} &\textbf{86.83} & \textbf{69.02} \\
      \bottomrule
  
  \end{tabular}
  }
\end{table}

\textbf{Experiments with the MCXFace Dataset:} Table~\ref{tab:mcxface} presents the average performance over five folds for the VIS-Thermal protocol on the MCXFace dataset, with results reported as mean and standard deviation across all folds. The baseline shows the performance of the pretrained \textit{IresNet100} face recognition model when evaluated directly on thermal images. As shown in the table, the proposed \textbf{xEdgeFace} method surpasses all competing approaches, achieving the highest average Rank-1 accuracy of 91.68\%.

\begin{table}[h]
  \caption{Performance of the proposed approach in the VIS-Thermal protocol of  MCXFace dataset, the Baseline is a pretrained \textit{Iresnet100} model.}
  \label{tab:mcxface}
  \centering
  \resizebox{0.98\columnwidth}{!}{%
  \begin{tabular}{lrrr}
  \toprule
  \textbf{Method} & \textbf{AUC}   & \textbf{EER}   & \textbf{Rank-1}   \\ \midrule
 Baseline & 84.45 $\pm$ 3.70  & 22.07 $\pm$ 2.81 & 47.23 $\pm$ 3.93    \\
DSU-Iresnet100 \cite{george2022prepended} & 98.12 $\pm$ 0.75 & 6.58 $\pm$ 1.35 & 83.43 $\pm$ 5.47 \\
PDT \cite{george2022prepended}      & 98.43 $\pm$ 0.78  &  6.52 $\pm$ 1.45  & 84.52 $\pm$ 5.36   \\ 
CAIM \cite{george2023bridging}  & 98.97 $\pm$ 0.24& 5.05 $\pm$ 0.91& 87.24$\pm$2.75\\\midrule
\rowcolor{Gray}
\textbf{xEdgeFace-Base (LN, ST)} &  99.12$\pm$0.14 &  4.71$\pm$0.31 &  89.98$\pm$2.47 \\

\rowcolor{Gray}
\textbf{xEdgeFace-Base (LN, ST, S0)} &  99.50$\pm$0.21 &  3.42$\pm$0.78 &  \textbf{91.68$\pm$2.67}\\
\bottomrule
  \end{tabular}
  }
  \end{table}

\textbf{Modality agnostic performance on MCXFace dataset:} We further evaluated the baseline methods and the proposed approach on the MCXFace dataset under the ``VIS-UNIVERSAL'' protocols (using protocols from \cite{george2024modality}), where enrollment samples come from the VIS domain and probe samples may belong to any of the other modalities (Thermal, Near-Infrared, or Shortwave Infrared). Each protocol includes a training set that contains identities in both the source and target modalities, as well as a development (dev) set in which the source images are used for enrollment and the target images are used for probing. Model training and selection are performed exclusively on the training set, while the dev set is used only for performance comparison and not for additional tuning. For evaluation, experiments are conducted across all five protocol splits, and we report the mean and standard deviation of the results (we follow the same evaluation scheme reported in \cite{george2024modality}). 

Table~\ref{tab:mcxface_combined} presents the aggregated results across all probe modalities. The proposed approach achieves performance comparable to existing methods while operating with a substantially smaller computational footprint. Across the three probe modalities: Near-Infrared, Thermal, and Shortwave Infrared, our method attains an average verification rate of 94.11\%. These results demonstrate that our approach enables modality-agnostic heterogeneous matching with a single unified model, greatly enhancing flexibility by supporting both homogeneous and heterogeneous recognition, as well as potential cross-modal matching across different spectral domains.

\begin{table*}[htb!]
  \centering
  \caption{Experimental results on VIS-UNIVERSAL protocol of the MCXFace dataset-- aggregated performance.}
  \label{tab:mcxface_combined}
  \resizebox{0.7\textwidth}{!}{

\begin{tabular}{cccccc}
\toprule
\textbf{Modality} &             \textbf{AUC} &           \textbf{EER} &              \textbf{Rank-1} &         \textbf{VR@FAR=0.1$\%$} &            \textbf{VR@FAR=1$\%$} \\
\midrule
DSU \cite{george2022prepended} &  95.57$\pm$0.80 & 10.24$\pm$0.88 &  84.21$\pm$0.94 &  67.89$\pm$1.05 &  78.13$\pm$1.20 \\
PDT \cite{george2022prepended} & 96.16$\pm$1.60 &  9.60$\pm$2.07 &  80.90$\pm$2.49 &  64.63$\pm$5.87 &  76.30$\pm$2.49 \\
CAIM \cite{george2023bridging} &  99.45$\pm$0.12 & 3.67$\pm$0.33 &  90.92$\pm$1.30 &  79.64$\pm$2.46 &  91.58$\pm$0.68 \\ \midrule 
SSMB \cite{george2024modality} &  99.70$\pm$0.08 & \textbf{2.59$\pm$0.28} &  92.80$\pm$0.71 &  84.04$\pm$1.71 &  \textbf{94.50$\pm$1.44} \\
\rowcolor{Gray}
\textbf{xEdgeFace-Base (LN, ST, S0)} &  \textbf{99.72$\pm$0.07} & 2.76$\pm$0.42 &  \textbf{94.42$\pm$1.42} &  \textbf{85.59$\pm$2.10} &  94.11$\pm$1.24 \\

\bottomrule
\end{tabular}

  }

\end{table*}

\textbf{Experiments with the Polathermal Dataset:} We evaluate our method on the thermal-to-visible face recognition protocols of the Polathermal dataset, with results summarized in Table~\ref{tab:polathermal}. The reported values correspond to the average Rank-1 identification accuracy over the five protocols defined in~\cite{de2018heterogeneous}. As shown, the proposed \textbf{xEdgeFace} model achieves the highest performance, reaching an average Rank-1 accuracy of 97.31\%.

\begin{table}[ht]
\caption{Pola Thermal - Average Rank-1 recognition rate.}
\label{tab:polathermal}
\begin{center}
  \resizebox{0.7\columnwidth}{!}{
  \begin{tabular}{lr}
    \toprule
    \textbf{Method} & \textbf{Mean (Std. Dev.)} \\ \midrule
    
    DPM in \cite{hu2016polarimetric}   & 75.31 \% (-) \\ 
    CpNN in \cite{hu2016polarimetric}  & 78.72 \% (-) \\ 
    PLS in \cite{hu2016polarimetric}   & 53.05\% (-)  \\  \midrule

    LBPs + DoG in \cite{liao2009heterogeneous} & 36.8\% (3.5) \\ 
    ISV in \cite{de2016heterogeneous}       & 23.5\% (1.1) \\ 
    GFK in \cite{sequeira2017cross}             & 34.1\% (2.9) \\ 

    DSU(Best Result) \cite{de2018heterogeneous} & 76.3\% (2.1) \\

    DSU-Iresnet100 \cite{george2022prepended} & 88.2\% (5.8) \\
    PDT \cite{george2022prepended} & 97.1\% (1.3) \\ 
    
    CAIM \cite{george2023bridging} & 95.00\% (1.63) \\ \midrule
\rowcolor{Gray}
    \textbf{xEdgeFace-Base (LN, ST)} & 95.98\% (1.90) \\
    \rowcolor{Gray}
    \textbf{xEdgeFace-Base (LN, ST, S0)} & \textbf{97.31\% (1.96)} \\
    \bottomrule 
  \end{tabular}
  }
\end{center}
\end{table}

\textbf{Experiments with the SCFace Dataset:} The SCFace dataset introduces a challenging form of heterogeneity due to the quality differences between the gallery images (high-resolution mugshots) and the probe images (low-resolution surveillance camera images). Table~\ref{tab:scface} reports the results for the ``far'' protocol, which is the most difficult among the dataset’s evaluation protocols. As shown, our method achieves the best Rank-1 accuracy of 96.36\%. This demonstrates that our framework can effectively handle not only cross-modal variation but also substantial image quality degradation.

\begin{table}[h]
  \caption{Performance of the proposed approach in the SCFace dataset, performance reported on the \textit{far} protocol.}
  \label{tab:scface}
  \centering
  \resizebox{0.85\columnwidth}{!}{%
  \begin{tabular}{lrrrr}
  \toprule
               \textbf{Method} & \textbf{AUC}   & \textbf{EER}   & \textbf{Rank-1}    & \begin{tabular}[c]{@{}c@{}} \end{tabular} \\ \midrule

                                     DSU-Iresnet100 \cite{george2022prepended} & 97.18 & 8.37 & 79.53 \\

                              PDT \cite{george2022prepended}   &  98.31 & 6.98 &  84.19            \\ 
                               CAIM \cite{george2023bridging} &   98.81 &   5.09 &   86.05  \\

                             SSMB \cite{george2024modality} (N=1) & 98.77 &  5.91 &  87.73 \\
                             SSMB \cite{george2024modality} (N=2) & 98.67 & 6.36 & 86.82 \\ \midrule

                             \rowcolor{Gray}

                              \textbf{xEdgeFace-Base (LN, ST)}        &  \textbf{99.86} &  \textbf{1.82} &  \textbf{96.36} \\
                             \rowcolor{Gray}
                              \textbf{xEdgeFace-Base (LN, ST, S0)} &  99.71 &  2.27 &  93.18 \\
                
  \bottomrule
  \end{tabular}
  }
  \vspace{-2mm}

  \end{table}

\textbf{Experiments with the CUFSF Dataset:} We next evaluate our method on the challenging sketch-to-photo face recognition task. Table~\ref{tab:cufsf} reports the Rank-1 accuracies of other approaches under the evaluation protocol defined in~\cite{fang2020identity}. Our method achieves a Rank-1 accuracy of 80.30\%, ranking second only to SSMB~\cite{george2024modality}. Despite this strong relative performance, the absolute accuracy remains lower than in other modalities such as thermal or near-infrared, showing the difficulty of this VIS-Sketch matching task.
The CUFSF dataset contains viewed hand-drawn sketches~\cite{klum2014facesketchid}, which, although recognizable to humans, often omit or exaggerate key discriminative facial cues used by recognition models. Unlike other sensing modalities, sketches are heavily influenced by artistic style and interpretation, introducing a substantial domain gap. Nevertheless, our approach demonstrates competitive performance under these extreme cross-modal conditions, highlighting its robustness and adaptability.

\begin{table}[ht]
\caption{CUFSF: Rank-1 recognition rate in sketch to photo recognition.}
\label{tab:cufsf}

\begin{center}
\resizebox{0.49\columnwidth}{!}{%
  \begin{tabular}{lrr}
    \toprule
    \textbf{Method} & \textbf{Rank-1} \\ \midrule
    IACycleGAN \cite{fang2020identity} &64.94 \\
    DSU-Iresnet100 \cite{george2022prepended} & 67.06 \\ 
    PDT \cite{george2022prepended}  & 71.08 \\ 
    CAIM \cite{george2023bridging} & 76.38 \\ 
    SSMB \cite{george2024modality} (N=1) &  81.14\\
    SSMB \cite{george2024modality} (N=2) & \textbf{81.67} \\ \midrule
\rowcolor{Gray}
    \textbf{xEdgeFace-Base (LN, ST)} & 80.30 \\
\rowcolor{Gray}
    \textbf{xEdgeFace-Base (LN, ST, S0)} & 78.81 \\
    \bottomrule 
  \end{tabular}
  }
\end{center}
\vspace{-2mm}
\end{table}

\textbf{Experiments with the CASIA NIR-VIS 2.0 Dataset:} We further evaluate the proposed method on the CASIA NIR-VIS~2.0 dataset to examine its performance on the VIS-NIR matching task. Due to the relatively small domain gap, VIS-pretrained models already achieve strong baseline performance. To ensure more rigorous comparison, we adopt stricter evaluation criteria, reporting VR@FAR=0.1\% and VR@FAR=0.01\%. Following the standard protocol, results are presented as the average and standard deviation over 10 folds. As shown in Table~\ref{tab:casia}, our approach consistently surpasses existing state-of-the-art methods, demonstrating excellent performance in the VIS-NIR setting.

\begin{table}[h]
  \centering
  \caption{Experimental results on CASIA NIR-VIS 2.0.}
  \label{tab:casia}
  \resizebox{0.47\textwidth}{!}{
  \begin{tabular}{l|ccc}
    \toprule
      \textbf{Method} & \textbf{Rank-1} & \textbf{VR@FAR=0.1$\%$} & \textbf{VR@FAR=0.01$\%$} \\
      \midrule
      IDNet \cite{Reale2016SeeingTF} & 87.1$\pm$0.9 & 74.5 & - \\
      HFR-CNN \cite{saxena2016heterogeneous} & 85.9$\pm$0.9 & 78.0 & - \\
      Hallucination \cite{Lezama2017NotAO} & 89.6$\pm$0.9 & - & - \\
      TRIVET \cite{XXLiu:2016} & 95.7$\pm$0.5 & 91.0$\pm$1.3 & 74.5$\pm$0.7 \\
      W-CNN \cite{DBLP:journals/corr/abs-1708-02412} & 98.7$\pm$0.3 & 98.4$\pm$0.4 & 94.3$\pm$0.4 \\
      PACH \cite{duan2019pose} & 98.9$\pm$0.2 & 98.3$\pm$0.2 & - \\
      RCN \cite{deng2019residual} & 99.3$\pm$0.2 & 98.7$\pm$0.2 & - \\
      MC-CNN \cite{8624555} & 99.4$\pm$0.1 & 99.3$\pm$0.1 & - \\
      DVR \cite{XWu:2019}  & 99.7$\pm$0.1 & 99.6$\pm$0.3 & 98.6$\pm$0.3 \\
      DVG \cite{fu2019dual} & 99.8$\pm$0.1 & 99.8$\pm$0.1 & 98.8$\pm$0.2 \\
      DVG-Face \cite{fu2021dvg} & 99.9$\pm$0.1 & 99.9$\pm$0.0 & 99.2$\pm$0.1 \\     
      PDT \cite{george2022prepended} & 99.95$\pm$0.04 & 99.94$\pm$0.03 & 99.77$\pm$0.09 \\ 
      CAIM \cite{george2023bridging} & 99.96$\pm$0.02 &\textbf{99.95$\pm$0.02} & 99.79$\pm$0.11 \\ \midrule
      \rowcolor{Gray}
      \textbf{xEdgeFace-Base (LN, ST)} & 99.96$\pm$0.02 & 99.91$\pm$0.02 & 99.83$\pm$0.04 \\
      \rowcolor{Gray}
      \textbf{xEdgeFace-Base (LN, ST, S0)} & \textbf{99.99$\pm$0.01} & 99.93$\pm$0.02 & \textbf{99.86$\pm$0.04}\\
      \bottomrule
  \end{tabular}}
\end{table}

\vspace{0mm}

\subsection{Discussions}
The comprehensive experimental results across six heterogeneous face recognition (HFR) benchmarks highlight the effectiveness, generalizability, and efficiency of the proposed \emph{xEdgeFace} framework. Despite its lightweight design, xEdgeFace consistently outperforms or closely matches state-of-the-art methods across a wide range of modalities, including thermal, NIR, sketch, and low-resolution surveillance imagery. Notably, our approach achieves top Rank-1 accuracy on several challenging datasets with minimal degradation in performance on standard face recognition benchmarks. The ablation studies provide valuable insights into the most impactful components of the adaptation strategy and demonstrate the crucial role of self-distillation in balancing cross-modal alignment while preventing catastrophic forgetting. The framework also scales effectively to highly compact model variants, achieving large relative improvements showcasing its suitability for edge and resource-constrained scenarios. Moreover, strong performance under large domain shifts (e.g., visible to thermal) further validates the robustness and adaptability of the proposed method across challenging heterogeneous settings. These results suggest that heterogeneous face recognition can be effectively addressed through parameter-efficient adaptation of pretrained RGB models, without requiring heavy architectures or modality-specific branches.

\subsection{Limitations}

The proposed framework is most effective for modality gaps dominated by spectral or photometric differences, such as VIS--NIR and VIS--Thermal, where adapting LayerNorm parameters and shallow layers can reduce domain gap while preserving RGB recognition performance. Its effectiveness is more limited for modalities with stronger structural discrepancies, such as sketch-to-photo matching, since these differences cannot be fully addressed through statistical alignment alone. This is reflected in the lower performance on CUFSF compared to the thermal and NIR benchmarks. More generally, our method should be viewed as a lightweight and parameter-efficient adaptation strategy that is particularly well suited to cross-spectral face recognition, rather than a universal solution for all heterogeneous settings. For modalities with larger structural gaps, more expressive adaptation mechanisms may be required.

\section{Conclusions}

In this work, we presented \emph{xEdgeFace}, an efficient framework for lightweight heterogeneous face recognition (HFR) that extends existing face recognition models to cross-modal scenarios. By selectively adapting early convolutional layers and LayerNorm (LN) modules within a contrastive self-distillation framework, our approach achieves strong cross-modal generalization while preserving the model’s original performance in the visible spectrum. This design enables the adapted model to perform robustly across challenging HFR tasks, including VIS-Thermal, VIS-NIR, and sketch-to-photo recognition while maintaining competitive accuracy on standard FR benchmarks, effectively mitigating catastrophic forgetting. Extensive experiments demonstrate that xEdgeFace consistently outperforms or matches state-of-the-art methods, even when applied to highly compact model variants, making it particularly suitable for edge deployment. The proposed model achieved performance comparable to, or better than, state-of-the-art models while using roughly one-twentieth of the compute. Furthermore, the results show that our adaptation strategy supports modality-agnostic heterogeneous matching within a single unified architecture, greatly enhancing flexibility by enabling both homogeneous and heterogeneous recognition, as well as potential cross-modal matching across different spectral domains. The source code and pretrained models are made publicly available to support reproducibility and facilitate further extensions of this work.

\ifCLASSOPTIONcompsoc
  \section*{Acknowledgments}
\else
  \section*{Acknowledgment}
\fi

This research was funded by the European Union project CarMen (Grant Agreement No. 101168325).

\ifCLASSOPTIONcaptionsoff
  \newpage
\fi



%
\bibliographystyle{IEEEtran}

\bibliography{egbib}

@article{hu2024pseudo,
  title={Pseudo label association and prototype-based invariant learning for semi-supervised nir-vis face recognition},
  author={Hu, Weipeng and Yang, Yiming and Hu, Haifeng},
  journal={IEEE Transactions on Image Processing},
  volume={33},
  pages={1448--1463},
  year={2024},
  publisher={IEEE}
}

@article{yang2023robust,
  title={Robust cross-domain pseudo-labeling and contrastive learning for unsupervised domain adaptation NIR-VIS face recognition},
  author={Yang, Yiming and Hu, Weipeng and Lin, Haiqi and Hu, Haifeng},
  journal={IEEE Transactions on Image Processing},
  volume={32},
  pages={5231--5244},
  year={2023},
  publisher={IEEE}
}

@article{yang2023unsupervised,
  title={Unsupervised NIR-VIS face recognition via homogeneous-to-heterogeneous learning and residual-invariant enhancement},
  author={Yang, Yiming and Hu, Weipeng and Hu, Haifeng},
  journal={IEEE Transactions on Information Forensics and Security},
  volume={19},
  pages={2112--2126},
  year={2023},
  publisher={IEEE}
}

@inproceedings{de2025towards,
  title={Towards Robust Facial Recognition: Gabor Filter-Based Feature Extraction for NIR-VIS Heterogeneous Face Recognition},
  author={de Andrade, Jo{\~a}o VR and Freire, Agostinho and Pereira, Guilherme Carvalho and Millan-Arias, Cristian and Fernandes, Bruno and Bastos-Filho, Carmelo and Tortato, Jorge and Da Rocha, Luiz and Maciel, Alexandre MA},
  booktitle={Proceedings of the 40th ACM/SIGAPP Symposium on Applied Computing},
  pages={1275--1281},
  year={2025}
}

@article{ba2016layer,
  title={Layer normalization},
  author={Ba, Jimmy Lei and Kiros, Jamie Ryan and Hinton, Geoffrey E},
  journal={arXiv preprint arXiv:1607.06450},
  year={2016}
}

@article{xedgeface,
  title     = {{xEdgeFace}: Efficient Cross-Spectral Face Recognition for Edge Devices},
  author    = {George, Anjith and Marcel, Sebastien},
  booktitle = {2025 IEEE International Joint Conference on Biometrics (IJCB)},
  pages     = {1--10},
  year      = {2025},
  organization = {IEEE}
}

@article{mostaani2020high,
  title={The high-quality wide multi-channel attack (HQ-WMCA) database},
  author={Mostaani, Zohreh and George, Anjith and Heusch, Guillaume and Geissbuhler, David and Marcel, Sebastien},
  journal={arXiv preprint arXiv:2009.09703},
  year={2020}
}

@article{xu2019understanding,
  title={Understanding and improving layer normalization},
  author={Xu, Jingjing and Sun, Xu and Zhang, Zhiyuan and Zhao, Guangxiang and Lin, Junyang},
  journal={Advances in neural information processing systems},
  volume={32},
  year={2019}
}

@inproceedings{george2025digi2real,
  title={Digi2real: Bridging the realism gap in synthetic data face recognition via foundation models},
  author={George, Anjith and Marcel, Sebastien},
  booktitle={Proceedings of the Winter Conference on Applications of Computer Vision},
  pages={1469--1478},
  year={2025}
}

@inproceedings{
zhao2024tuning,
title={Tuning LayerNorm in Attention: Towards Efficient Multi-Modal {LLM} Finetuning},
author={Bingchen Zhao and Haoqin Tu and Chen Wei and Jieru Mei and Cihang Xie},
booktitle={The Twelfth International Conference on Learning Representations},
year={2024},
url={https://openreview.net/forum?id=YR3ETaElNK}
}

@article{howard2017mobilenets,
  title={Mobilenets: Efficient convolutional neural networks for mobile vision applications},
  author={Howard, Andrew G and Zhu, Menglong and Chen, Bo and Kalenichenko, Dmitry and Wang, Weijun and Weyand, Tobias and Andreetto, Marco and Adam, Hartwig},
  journal={arXiv preprint arXiv:1704.04861},
  year={2017}
}

@inproceedings{sandler2018mobilenetv2,
  title={Mobilenetv2: Inverted residuals and linear bottlenecks},
  author={Sandler, Mark and Howard, Andrew and Zhu, Menglong and Zhmoginov, Andrey and Chen, Liang-Chieh},
  booktitle={Proceedings of the IEEE conference on computer vision and pattern recognition},
  pages={4510--4520},
  year={2018}
}

@inproceedings{chen2018mobilefacenets,
  title={Mobilefacenets: Efficient cnns for accurate real-time face verification on mobile devices},
  author={Chen, Sheng and Liu, Yang and Gao, Xiang and Han, Zhen},
  booktitle={Biometric Recognition: 13th Chinese Conference, CCBR 2018, Urumqi, China, August 11-12, 2018, Proceedings 13},
  pages={428--438},
  year={2018},
  organization={Springer}
}

@article{tan2019mixconv,
  title={Mixconv: Mixed depthwise convolutional kernels},
  author={Tan, Mingxing and Le, Quoc V},
  journal={arXiv preprint arXiv:1907.09595},
  year={2019}
}

@inproceedings{boutros2021mixfacenets,
  title={Mixfacenets: Extremely efficient face recognition networks},
  author={Boutros, Fadi and Damer, Naser and Fang, Meiling and Kirchbuchner, Florian and Kuijper, Arjan},
  booktitle={2021 IEEE International Joint Conference on Biometrics (IJCB)},
  pages={1--8},
  year={2021},
  organization={IEEE}
}

@inproceedings{wu2018shift,
  title={Shift: A zero flop, zero parameter alternative to spatial convolutions},
  author={Wu, Bichen and Wan, Alvin and Yue, Xiangyu and Jin, Peter and Zhao, Sicheng and Golmant, Noah and Gholaminejad, Amir and Gonzalez, Joseph and Keutzer, Kurt},
  booktitle={Proceedings of the IEEE conference on computer vision and pattern recognition},
  pages={9127--9135},
  year={2018}
}

@inproceedings{ma2018shufflenet,
  title={Shufflenet v2: Practical guidelines for efficient cnn architecture design},
  author={Ma, Ningning and Zhang, Xiangyu and Zheng, Hai-Tao and Sun, Jian},
  booktitle={Proceedings of the European conference on computer vision (ECCV)},
  pages={116--131},
  year={2018}
}

@inproceedings{martindez2019shufflefacenet,
  title={Shufflefacenet: A lightweight face architecture for efficient and highly-accurate face recognition},
  author={Martindez-Diaz, Yoanna and Luevano, Luis S and Mendez-Vazquez, Heydi and Nicolas-Diaz, Miguel and Chang, Leonardo and Gonzalez-Mendoza, Miguel},
  booktitle={Proceedings of the IEEE/CVF International Conference on Computer Vision Workshops},
  pages={0--0},
  year={2019}
}

@article{boutros2022pocketnet,
  title={Pocketnet: Extreme lightweight face recognition network using neural architecture search and multistep knowledge distillation},
  author={Boutros, Fadi and Siebke, Patrick and Klemt, Marcel and Damer, Naser and Kirchbuchner, Florian and Kuijper, Arjan},
  journal={IEEE Access},
  volume={10},
  pages={46823--46833},
  year={2022},
  publisher={IEEE}
}

@article{yi2014learning,
  title={Learning face representation from scratch},
  author={Yi, Dong and Lei, Zhen and Liao, Shengcai and Li, Stan Z},
  journal={arXiv preprint arXiv:1411.7923},
  year={2014}
}

@inproceedings{yan2019vargfacenet,
  title={Vargfacenet: An efficient variable group convolutional neural network for lightweight face recognition},
  author={Yan, Mengjia and Zhao, Mengao and Xu, Zining and Zhang, Qian and Wang, Guoli and Su, Zhizhong},
  booktitle={Proceedings of the IEEE/CVF International Conference on Computer Vision Workshops},
  pages={0--0},
  year={2019}
}

@inproceedings{deng2019lightweight,
  title={Lightweight face recognition challenge},
  author={Deng, Jiankang and Guo, Jia and Zhang, Debing and Deng, Yafeng and Lu, Xiangju and Shi, Song},
  booktitle={Proceedings of the IEEE/CVF International Conference on Computer Vision Workshops},
  pages={0--0},
  year={2019}
}

@article{alansari2023ghostfacenets,
  title={GhostFaceNets: Lightweight Face Recognition Model from Cheap Operations},
  author={Alansari, Mohamad and Hay, Oussama Abdul and Javed, Sajid and Shoufan, Abdulhaid and Zweiri, Yahya and Werghi, Naoufel},
  journal={IEEE Access},
  year={2023},
  publisher={IEEE}
}

@article{george2024edgeface,
  title={Edgeface: Efficient face recognition model for edge devices},
  author={George, Anjith and Ecabert, Christophe and Shahreza, Hatef Otroshi and Kotwal, Ketan and Marcel, S{\'e}bastien},
  journal={IEEE Transactions on Biometrics, Behavior, and Identity Science},
  volume={6},
  number={2},
  pages={158--168},
  year={2024},
  publisher={IEEE}
}

@inproceedings{george2024modality,
  title={Modality Agnostic Heterogeneous Face Recognition with Switch Style Modulators},
  author={George, Anjith and Marcel, S{\'e}bastien},
  booktitle={2024 IEEE International Joint Conference on Biometrics (IJCB)},
  pages={1--10},
  year={2024},
  organization={IEEE}
}

@article{klum2014facesketchid,
  title={The FaceSketchID system: Matching facial composites to mugshots},
  author={Klum, Scott J and Han, Hu and Klare, Brendan F and Jain, Anil K},
  journal={IEEE Transactions on Information Forensics and Security},
  volume={9},
  number={12},
  pages={2248--2263},
  year={2014},
  publisher={IEEE}
}

@inproceedings{gatys2016image,
  title={Image style transfer using convolutional neural networks},
  author={Gatys, Leon A and Ecker, Alexander S and Bethge, Matthias},
  booktitle={Proceedings of the IEEE conference on computer vision and pattern recognition},
  pages={2414--2423},
  year={2016}
}

@article{george2024modalities,
  title={From modalities to styles: Rethinking the domain gap in heterogeneous face recognition},
  author={George, Anjith and Marcel, S{\'e}bastien},
  journal={IEEE Transactions on Biometrics, Behavior, and Identity Science},
  volume={6},
  number={4},
  pages={475--485},
  year={2024},
  publisher={IEEE}
}

@inproceedings{maaz2023edgenext,
  title={Edgenext: efficiently amalgamated cnn-transformer architecture for mobile vision applications},
  author={Maaz, Muhammad and Shaker, Abdelrahman and Cholakkal, Hisham and Khan, Salman and Zamir, Syed Waqas and Anwer, Rao Muhammad and Shahbaz Khan, Fahad},
  booktitle={Computer Vision--ECCV 2022 Workshops: Tel Aviv, Israel, October 23--27, 2022, Proceedings, Part VII},
  pages={3--20},
  year={2023},
  organization={Springer}
}

@inproceedings{george2024heterogeneous,
  title={Heterogeneous face recognition using domain invariant units},
  author={George, Anjith and Marcel, S{\'e}bastien},
  booktitle={ICASSP 2024-2024 IEEE International Conference on Acoustics, Speech and Signal Processing (ICASSP)},
  pages={4780--4784},
  year={2024},
  organization={IEEE}
}

@inproceedings{liu2023modality,
  title={Modality-agnostic Augmented Multi-Collaboration Representation for Semi-supervised Heterogenous Face Recognition},
  author={Liu, Decheng and Yang, Weizhao and Peng, Chunlei and Wang, Nannan and Hu, Ruimin and Gao, Xinbo},
  booktitle={Proceedings of the 31st ACM International Conference on Multimedia},
  pages={4647--4656},
  year={2023}
}

@article{han2020model,
  title={Model rubik’s cube: Twisting resolution, depth and width for tinynets},
  author={Han, Kai and Wang, Yunhe and Zhang, Qiulin and Zhang, Wei and Xu, Chunjing and Zhang, Tong},
  journal={Advances in Neural Information Processing Systems},
  volume={33},
  pages={19353--19364},
  year={2020}
}

@inproceedings{huang2008labeled,
  title={Labeled faces in the wild: A database forstudying face recognition in unconstrained environments},
  author={Huang, Gary B and Mattar, Marwan and Berg, Tamara and Learned-Miller, Eric},
  booktitle={Workshop on faces in'Real-Life'Images: detection, alignment, and recognition},
  year={2008}
}

@inproceedings{sengupta2016frontal,
  title={Frontal to profile face verification in the wild},
  author={Sengupta, Soumyadip and Chen, Jun-Cheng and Castillo, Carlos and Patel, Vishal M and Chellappa, Rama and Jacobs, David W},
  booktitle={2016 IEEE winter conference on applications of computer vision (WACV)},
  pages={1--9},
  year={2016},
  organization={IEEE}
}

@inproceedings{moschoglou2017agedb,
  title={Agedb: the first manually collected, in-the-wild age database},
  author={Moschoglou, Stylianos and Papaioannou, Athanasios and Sagonas, Christos and Deng, Jiankang and Kotsia, Irene and Zafeiriou, Stefanos},
  booktitle={proceedings of the IEEE conference on computer vision and pattern recognition workshops},
  pages={51--59},
  year={2017}
}

@article{hu2022lora,
  title={Lora: Low-rank adaptation of large language models.},
  author={Hu, Edward J and Shen, Yelong and Wallis, Phillip and Allen-Zhu, Zeyuan and Li, Yuanzhi and Wang, Shean and Wang, Lu and Chen, Weizhu and others},
  journal={ICLR},
  volume={1},
  number={2},
  pages={3},
  year={2022}
}

@article{zheng2017cross,
  title={Cross-age lfw: A database for studying cross-age face recognition in unconstrained environments},
  author={Zheng, Tianyue and Deng, Weihong and Hu, Jiani},
  journal={arXiv preprint arXiv:1708.08197},
  year={2017}
}

@article{zheng2018cross,
  title={Cross-pose lfw: A database for studying cross-pose face recognition in unconstrained environments},
  author={Zheng, Tianyue and Deng, Weihong},
  journal={Beijing University of Posts and Telecommunications, Tech. Rep},
  volume={5},
  number={7},
  year={2018}
}

@article{shahreza2024knowledge,
  title={Knowledge distillation for face recognition using synthetic data with dynamic latent sampling},
  author={Shahreza, Hatef Otroshi and George, Anjith and Marcel, S{\'e}bastien},
  journal={IEEE Access},
  year={2024},
  publisher={IEEE}
}

@article{khan2022transformers,
  title={Transformers in vision: A survey},
  author={Khan, Salman and Naseer, Muzammal and Hayat, Munawar and Zamir, Syed Waqas and Khan, Fahad Shahbaz and Shah, Mubarak},
  journal={ACM computing surveys (CSUR)},
  volume={54},
  number={10s},
  pages={1--41},
  year={2022},
  publisher={ACM New York, NY}
}

@article{liu2021heterogeneous,
  title={Heterogeneous face interpretable disentangled representation for joint face recognition and synthesis},
  author={Liu, Decheng and Gao, Xinbo and Peng, Chunlei and Wang, Nannan and Li, Jie},
  journal={IEEE transactions on neural networks and learning systems},
  volume={33},
  number={10},
  pages={5611--5625},
  year={2021},
  publisher={IEEE}
}

@article{liu2020coupled,
  title={Coupled attribute learning for heterogeneous face recognition},
  author={Liu, Decheng and Gao, Xinbo and Wang, Nannan and Li, Jie and Peng, Chunlei},
  journal={IEEE Transactions on Neural Networks and Learning Systems},
  volume={31},
  number={11},
  pages={4699--4712},
  year={2020},
  publisher={IEEE}
}

@article{liu2018composite,
  title={Composite components-based face sketch recognition},
  author={Liu, Decheng and Li, Jie and Wang, Nannan and Peng, Chunlei and Gao, Xinbo},
  journal={Neurocomputing},
  volume={302},
  pages={46--54},
  year={2018},
  publisher={Elsevier}
}

@inproceedings{george2023bridging,
  title={{Bridging the Gap}: Heterogeneous Face Recognition with Conditional Adaptive Instance Modulation},
  author={George, Anjith and Marcel, Sebastien},
booktitle={2023 International Joint Conference on Biometrics (IJCB)},
organization={IEEE},
year={2023}
}

@inproceedings{zhu2021webface260m,
  title={Webface260m: A benchmark unveiling the power of million-scale deep face recognition},
  author={Zhu, Zheng and Huang, Guan and Deng, Jiankang and Ye, Yun and Huang, Junjie and Chen, Xinze and Zhu, Jiagang and Yang, Tian and Lu, Jiwen and Du, Dalong and others},
  booktitle={Proceedings of the IEEE/CVF Conference on Computer Vision and Pattern Recognition},
  pages={10492--10502},
  year={2021}
}

@article{luo2022memory,
  title={Memory-Modulated Transformer Network for Heterogeneous Face Recognition},
  author={Luo, Mandi and Wu, Haoxue and Huang, Huaibo and He, Weizan and He, Ran},
  journal={IEEE Transactions on Information Forensics and Security},
  year={2022},
  publisher={IEEE}
}

@article{phillips1998feret,
  title={The {FERET} database and evaluation procedure for face-recognition algorithms},
  author={Phillips, P Jonathon and Wechsler, Harry and Huang, Jeffery and Rauss, Patrick J},
  journal={Image and vision computing},
  volume={16},
  number={5},
  pages={295--306},
  year={1998},
  publisher={Elsevier}
}

@article{fang2020identity,
  title={{Identity-aware {CycleGAN} for face photo-sketch synthesis and recognition}},
  author={Fang, Yuke and Deng, Weihong and Du, Junping and Hu, Jiani},
  journal={Pattern Recognition},
  volume={102},
  pages={107249},
  year={2020},
  publisher={Elsevier}
}

@inproceedings{zhang2011coupled,
  title={Coupled information-theoretic encoding for face photo-sketch recognition},
  author={Zhang, Wei and Wang, Xiaogang and Tang, Xiaoou},
  booktitle={CVPR 2011},
  pages={513--520},
  year={2011},
  organization={IEEE}
}

@article{li2007illumination,
  title={Illumination invariant face recognition using near-infrared images},
  author={Li, Stan Z and Chu, RuFeng and Liao, ShengCai and Zhang, Lun},
  journal={IEEE Transactions on pattern analysis and machine intelligence},
  volume={29},
  number={4},
  pages={627--639},
  year={2007},
  publisher={IEEE}
}

@article{he2018wasserstein,
  title={Wasserstein {CNN}: Learning invariant features for {Nir-Vis} face recognition},
  author={He, Ran and Wu, Xiang and Sun, Zhenan and Tan, Tieniu},
  journal={IEEE transactions on pattern analysis and machine intelligence},
  volume={41},
  number={7},
  pages={1761--1773},
  year={2018},
  publisher={IEEE}
}

@inproceedings{he2017learning,
  title={Learning invariant deep representation for {Nir-Vis} face recognition},
  author={He, Ran and Wu, Xiang and Sun, Zhenan and Tan, Tieniu},
  booktitle={Thirty-First AAAI Conference on Artificial Intelligence},
  year={2017}
}

@article{roy2018novel,
  title={A novel quaternary pattern of local maximum quotient for heterogeneous face recognition},
  author={Roy, Hiranmoy and Bhattacharjee, Debotosh},
  journal={Pattern Recognition Letters},
  volume={113},
  pages={19--28},
  year={2018},
  publisher={Elsevier}
}

@inproceedings{bob2017,
  author = {A. Anjos AND M. G\"unther AND T. de Freitas Pereira AND
            P. Korshunov AND A. Mohammadi AND S. Marcel},
  title = {Continuously Reproducing Toolchains in Pattern Recognition and
           Machine Learning Experiments},
  year = {2017},
  month = aug,
  booktitle = {International Conference on Machine Learning (ICML)},
  url = {http://publications.idiap.ch/downloads/papers/2017/Anjos_ICML2017-2_2017.pdf}
}

@article{klare2010matching,
  title={Matching forensic sketches to mug shot photos},
  author={Klare, Brendan and Li, Zhifeng and Jain, Anil K},
  journal={IEEE transactions on pattern analysis and machine intelligence},
  volume={33},
  number={3},
  pages={639--646},
  year={2010},
  publisher={IEEE}
}

@inproceedings{liao2009heterogeneous,
  title={Heterogeneous face recognition from local structures of normalized appearance},
  author={Liao, Shengcai and Yi, Dong and Lei, Zhen and Qin, Rui and Li, Stan Z},
  booktitle={International Conference on Biometrics},
  pages={209--218},
  year={2009},
  organization={Springer}
}

@article{klare2012heterogeneous,
  title={Heterogeneous face recognition using kernel prototype similarities},
  author={Klare, Brendan F and Jain, Anil K},
  journal={IEEE transactions on pattern analysis and machine intelligence},
  volume={35},
  number={6},
  pages={1410--1422},
  year={2012},
  publisher={IEEE}
}

@inproceedings{sharma2011bypassing,
  title={Bypassing synthesis: {PLS} for face recognition with pose, low-resolution and sketch},
  author={Sharma, Abhishek and Jacobs, David W},
  booktitle={CVPR 2011},
  pages={593--600},
  year={2011},
  organization={IEEE}
}

@inproceedings{lei2009coupled,
  title={Coupled spectral regression for matching heterogeneous faces},
  author={Lei, Zhen and Li, Stan Z},
  booktitle={2009 IEEE Conference on Computer Vision and Pattern Recognition},
  pages={1123--1128},
  year={2009},
  organization={IEEE}
}

@inproceedings{yi2007face,
  title={Face matching between near infrared and visible light images},
  author={Yi, Dong and Liu, Rong and Chu, RuFeng and Lei, Zhen and Li, Stan Z},
  booktitle={International Conference on Biometrics},
  pages={523--530},
  year={2007},
  organization={Springer}
}

@article{grgic2011scface,
  title={{SCface}--surveillance cameras face database},
  author={Grgic, Mislav and Delac, Kresimir and Grgic, Sonja},
  journal={Multimedia tools and applications},
  volume={51},
  number={3},
  pages={863--879},
  year={2011},
  publisher={Springer}
}

@article{learned2016labeled,
  title={Labeled faces in the wild: A survey},
  author={Learned-Miller, Erik and Huang, Gary B and RoyChowdhury, Aruni and Li, Haoxiang and Hua, Gang},
  journal={Advances in face detection and facial image analysis},
  pages={189--248},
  volume={1},
  year={2016},
  publisher={Springer}
}

@article{panetta2018comprehensive,
  title={A comprehensive database for benchmarking imaging systems},
  author={Panetta, Karen and Wan, Qianwen and Agaian, Sos and Rajeev, Srijith and Kamath, Shreyas and Rajendran, Rahul and Rao, Shishir Paramathma and Kaszowska, Aleksandra and Taylor, Holly A and Samani, Arash and others},
  journal={IEEE transactions on pattern analysis and machine intelligence},
  volume={42},
  number={3},
  pages={509--520},
  year={2018},
  publisher={IEEE}
}

@inproceedings{li2013casia,
  title={The {CASIA Nir-Vis 2.0} face database},
  author={Li, Stan and Yi, Dong and Lei, Zhen and Liao, Shengcai},
  booktitle={Proceedings of the IEEE conference on computer vision and pattern recognition workshops},
  pages={348--353},
  year={2013}
}

@inproceedings{zhang2017generative,
  title={Generative adversarial network-based synthesis of visible faces from polarimetric thermal faces},
  author={Zhang, He and Patel, Vishal M and Riggan, Benjamin S and Hu, Shuowen},
  booktitle={2017 IEEE International Joint Conference on Biometrics (IJCB)},
  pages={100--107},
  year={2017},
  organization={IEEE}
}

@inproceedings{liu2005nonlinear,
  title={A nonlinear approach for face sketch synthesis and recognition},
  author={Liu, Qingshan and Tang, Xiaoou and Jin, Hongliang and Lu, Hanqing and Ma, Songde},
  booktitle={2005 IEEE Computer Society conference on computer vision and pattern recognition (CVPR'05)},
  volume={1},
  pages={1005--1010},
  year={2005},
  organization={IEEE}
}

@article{wang2008face,
  title={Face photo-sketch synthesis and recognition},
  author={Wang, Xiaogang and Tang, Xiaoou},
  journal={IEEE transactions on pattern analysis and machine intelligence},
  volume={31},
  number={11},
  pages={1955--1967},
  year={2008},
  publisher={IEEE}
}

@article{fu2021dvg,
  title={{DVG-face}: Dual variational generation for heterogeneous face recognition},
  author={Fu, Chaoyou and Wu, Xiang and Hu, Yibo and Huang, Huaibo and He, Ran},
  journal={IEEE Transactions on Pattern Analysis and Machine Intelligence},
  year={2021},
  publisher={IEEE}
}

@article{Wu2018ALC,
  title={A light {CNN} for deep face representation with noisy labels},
  author={Wu, Xiang and He, Ran and Sun, Zhenan and Tan, Tieniu},
  journal={IEEE Transactions on Information Forensics and Security},
  volume={13},
  number={11},
  pages={2884--2896},
  year={2018}
}

@inproceedings{fu2019dual,
  title={Dual Variational Generation for Low Shot Heterogeneous Face Recognition},
  author={Fu, Chaoyou and Wu, Xiang and Hu, Yibo and Huang, Huaibo and He, Ran},
  booktitle={Advances in Neural Information Processing Systems},
  year={2019}
}

@article{de2018heterogeneous,
  title={Heterogeneous face recognition using domain specific units},
  author={de Freitas Pereira, Tiago and Anjos, Andr{\'e} and Marcel, S{\'e}bastien},
  journal={IEEE Transactions on Information Forensics and Security},
  volume={14},
  number={7},
  pages={1803--1816},
  year={2018},
  publisher={IEEE}
}

@inproceedings{hu2016polarimetric,
  title={A polarimetric thermal database for face recognition research},
  author={Hu, Shuowen and Short, Nathaniel J and Riggan, Benjamin S and Gordon, Christopher and Gurton, Kristan P and Thielke, Matthew and Gurram, Prudhvi and Chan, Alex L},
  booktitle={Proceedings of the IEEE conference on computer vision and pattern recognition workshops},
  pages={119--126},
  year={2016}
}

@inproceedings{de2016heterogeneous,
  title={Heterogeneous face recognition using inter-session variability modelling},
  author={de Freitas Pereira, Tiago and Marcel, S{\'e}bastien},
  booktitle={Proceedings of the IEEE Conference on Computer Vision and Pattern Recognition Workshops},
  pages={111--118},
  year={2016}
}

@inproceedings{sequeira2017cross,
  title={Cross-eyed 2017: Cross-spectral iris/periocular recognition competition},
  author={Sequeira, Ana F and Chen, Lulu and Ferryman, James and Wild, Peter and Alonso-Fernandez, Fernando and Bigun, Josef and Raja, Kiran B and Raghavendra, Ramachandra and Busch, Christoph and de Freitas Pereira, Tiago and others},
  booktitle={2017 IEEE International Joint Conference on Biometrics (IJCB)},
  pages={725--732},
  year={2017},
  organization={IEEE}
}

@inproceedings{deng2019residual,
  title={Residual compensation networks for heterogeneous face recognition},
  author={Deng, Zhongying and Peng, Xiaojiang and Qiao, Yu},
  booktitle={AAAI Conference on Artificial Intelligence},
  year={2019}
}

@article{8624555,
  title={Mutual component convolutional neural networks for heterogeneous face recognition},
  author={Deng, Zhongying and Peng, Xiaojiang and Li, Zhifeng and Qiao, Yu},
  journal={IEEE Transactions on Image Processing},
  volume={28},
  number={6},
  pages={3102--3114},
  year={2019}
}

@inproceedings{XWu:2019,
  title={Disentangled variational representation for heterogeneous face recognition},
  author={Wu, Xiang and Huang, Huaibo and Patel, Vishal M and He, Ran and Sun, Zhenan},
  booktitle={AAAI Conference on Artificial Intelligence},
  year={2019}
}

@inproceedings{duan2019pose,
  title={Pose Agnostic Cross-spectral Hallucination via Disentangling Independent Factors},
  author={Duan, Boyan and Fu, Chaoyou and Li, Yi and Song, Xingguang and He, Ran},
  booktitle={IEEE Conference on Computer Vision and Pattern Recognition},
  year={2020}
}

@inproceedings{XXLiu:2016,
  title={Transferring deep representation for {Nir-Vis} heterogeneous face recognition},
  author={Liu, Xiaoxiang and Song, Lingxiao and Wu, Xiang and Tan, Tieniu},
  booktitle={International Conference on Biometrics},
  year={2016}
}

@article{DBLP:journals/corr/abs-1708-02412,
  title={Wasserstein {CNN}: Learning invariant features for {Nir-Vis} face recognition},
  author={He, Ran and Wu, Xiang and Sun, Zhenan and Tan, Tieniu},
  journal={IEEE Transactions on Pattern Analysis and Machine Intelligence},
  volume={41},
  number={7},
  pages={1761--1773},
  year={2018}
}

@inproceedings{saxena2016heterogeneous,
  title={Heterogeneous face recognition with {CNNs}},
  author={Saxena, Shreyas and Verbeek, Jakob},
  booktitle={European Conference on Computer Vision},
  year={2016}
}

@inproceedings{Lezama2017NotAO,
  title={Not afraid of the dark: {Nir-Vis} face recognition via cross-spectral hallucination and low-rank embedding},
  author={Lezama, Jos{\'e} and Qiu, Qiang and Sapiro, Guillermo},
  booktitle={IEEE Conference on Computer Vision and Pattern Recognition},
  year={2017}
}

@inproceedings{Reale2016SeeingTF,
  title={Seeing the forest from the trees: A holistic approach to near-infrared heterogeneous face recognition},
  author={Reale, Christopher and Nasrabadi, Nasser M and Kwon, Heesung and Chellappa, Rama},
  booktitle={IEEE Conference on Computer Vision and Pattern Recognition Workshops},
  year={2016}
}

@article{george2022comprehensive,
  title={A Comprehensive Evaluation on Multi-channel Biometric Face Presentation Attack Detection},
  author={George, Anjith and Geissbuhler, David and Marcel, Sebastien},
  journal={arXiv preprint arXiv:2202.10286},
  year={2022}
}

@article{george2022prepended,
  title={Prepended Domain Transformer: Heterogeneous Face Recognition without Bells and Whistles},
  author={George, Anjith and Mohammadi, Amir and Marcel, Sebastien},
  journal={IEEE Transactions on Information Forensics and Security},
  year={2022},
  publisher={IEEE}
}

@article{zhuUnpairedImagetoImageTranslation2017,
  title = {Unpaired {{Image-to-Image Translation}} Using {{Cycle-Consistent Adversarial Networks}}},
  author = {Zhu, Jun-Yan and Park, Taesung and Isola, Phillip and Efros, Alexei A.},
  year = {2017},
  month = mar,
  journal = {arXiv:1703.10593 [cs]},
  eprint = {1703.10593},
  eprinttype = {arxiv},
  primaryclass = {cs},
  archiveprefix = {arXiv},
}

@inproceedings{kolf2023efar,
  title={{EFaR 2023}: Efficient Face Recognition Competition},
  author={Kolf, Jan Niklas and Boutros, Fadi and Elliesen, Jurek and Theuerkauf, Markus and Damer, Naser and Alansari, Mohamad and Hay, Oussama Abdul and Alansari, Sara and Javed, Sajid and Werghi, Naoufel and others},
booktitle={2023 International Joint Conference on Biometrics (IJCB)},
organization={IEEE},
year={2023}
}

@inproceedings{bob2012,
  author = {A. Anjos AND L. El Shafey AND R. Wallace AND
            M. G\"unther AND C. McCool AND S. Marcel},
  title = {Bob: a free signal processing and machine learning toolbox for researchers},
  year = {2012},
  month = oct,
  booktitle = {20th ACM Conference on Multimedia Systems (ACMMM), Nara, Japan},
  url = {https://publications.idiap.ch/downloads/papers/2012/Anjos_Bob_ACMMM12.pdf},
}

\begin{IEEEbiography}[{\includegraphics[width=1in,height=1.25in, trim={0cm 0.5cm 0cm 0.3cm},clip,keepaspectratio]{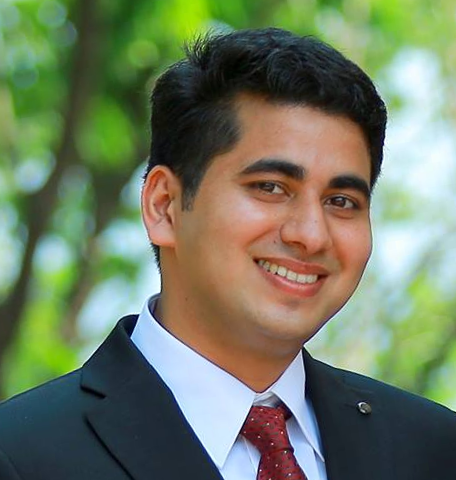}}]{Anjith George} has received his Ph.D. and M-Tech degree from the Department of Electrical Engineering, Indian Institute of Technology (IIT) Kharagpur, India in 2012 and 2018 respectively. After Ph.D, he worked in Samsung Research Institute as a machine learning researcher. Currently, he is a research associate in the biometric security and privacy group at Idiap Research Institute, focusing on developing face recognition and presentation attack detection algorithms. His research interests are real-time signal and image processing, embedded systems, computer vision, machine learning with a special focus on Biometrics.
\end{IEEEbiography}

\begin{IEEEbiography}[{\includegraphics[width=1in,height=1.25in,trim={7cm 0cm 7cm 0.5cm},clip,keepaspectratio]{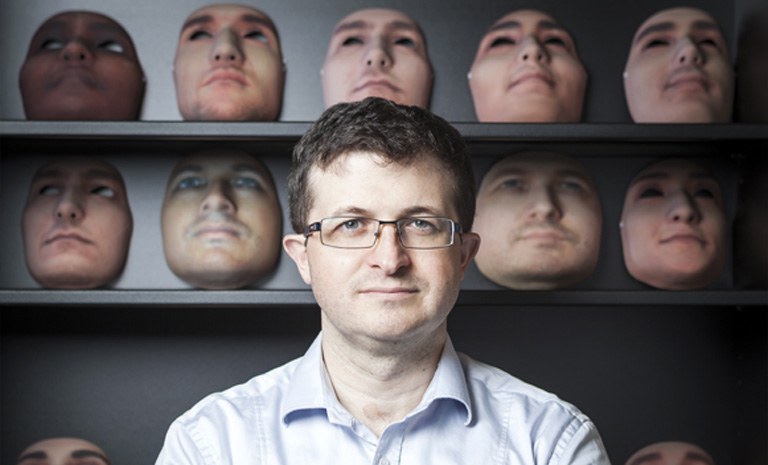}}]{S{\'e}bastien Marcel} heads the Biometrics Security and Privacy group at Idiap Research Institute (Switzerland) and conducts research on face recognition, speaker recognition, vein recognition, attack detection (presentation attacks, morphing attacks, deepfakes) and template protection. He received his Ph.D. degree in signal processing from Universit{\'e} de Rennes I in France (2000) at CNET, the research center of France Telecom (now Orange Labs). He is Professor at the University of Lausanne (School of Criminal Justice) and a lecturer at the  \'{E}cole Polytechnique F{\'e}d{\'e}rale de Lausanne. He is also the Director of the Swiss Center for Biometrics Research and Testing, which conducts certifications of biometric products.
\end{IEEEbiography}


\end{document}